%% file: main.tex
\documentclass[10pt,journal,compsoc]{IEEEtran}
\IEEEoverridecommandlockouts
\usepackage[table]{xcolor}
\usepackage{amsmath,amsfonts}
\usepackage{algorithmic}
\usepackage{algorithm}
\usepackage{array}
\usepackage[caption=false,font=normalsize,labelfont=sf,textfont=sf]{subfig}
\usepackage{textcomp}
\usepackage{stfloats}
\usepackage{verbatim}
\usepackage{graphicx}
\usepackage{cite}
\usepackage{times}
\usepackage{moreverb}
\usepackage{graphicx}
\usepackage{mathrsfs}
\usepackage{bm}
\usepackage{url}
\usepackage{amsmath}
\usepackage{amssymb}
\usepackage[edges]{forest}

\usepackage{graphicx}
\usepackage{amsmath,amssymb} 
\usepackage{times}
\usepackage{epsfig}
\usepackage{tabularx}
\usepackage{array}
\usepackage{booktabs}
\usepackage{amstext}
\usepackage{relsize}
\usepackage{lipsum}
\usepackage{color}

\usepackage{multirow}
\usepackage{enumitem}
\usepackage{makecell}
\usepackage{amsmath}
\usepackage{textcomp}
\usepackage{stfloats}
\usepackage{url}
\usepackage{verbatim}
\usepackage{graphicx}
\usepackage{tikz}
\usepackage[colorlinks, linkcolor=blue, citecolor=blue, urlcolor=blue]{hyperref}

\newcommand{\hide}[1]{}

\def\BibTeX{{\rm B\kern-.05em{\sc i\kern-.025em b}\kern-.08em
    T\kern-.1667em\lower.7ex\hbox{E}\kern-.125emX}}

\usepackage{balance}

\definecolor{hidden-draw}{RGB}{20,68,106}
\definecolor{hidden-pink}{RGB}{255,245,247}
\definecolor{lightred}{RGB}{255, 204, 204}
\definecolor{lightgreen}{RGB}{224, 255, 225}
\definecolor{lightyellow}{RGB}{255, 241, 224}
\definecolor{lightpurple}{RGB}{225, 225, 255}
\definecolor{lightgray}{gray}{0.9}

\title{A Survey of Generative Techniques for Spatial-Temporal Data Mining}

\author{Qianru Zhang\IEEEauthorrefmark{1}, Haixin Wang\IEEEauthorrefmark{1}, Cheng Long, Liangcai Su, Xingwei He, Jianlong Chang, Tailin Wu, Hongzhi Yin, Siu-Ming Yiu\IEEEauthorrefmark{2}, Qi Tian,~\IEEEmembership{Fellow, IEEE}, Christian S. Jensen,~\IEEEmembership{Fellow, IEEE}  
\IEEEcompsocitemizethanks{
\IEEEcompsocthanksitem
\IEEEauthorrefmark{1}Equal Contribution,\IEEEauthorrefmark{2}Corresponding author.
\IEEEcompsocthanksitem
Q. Zhang, X. He and S.M. Yiu are from the University of Hong Kong. E-mail: \{qrzhang,smyiu\}@cs.hku.hk, hexingwei15@gmail.com.
\IEEEcompsocthanksitem
H. Wang is from Peking University. E-mail: wang.hx@stu.pku.edu.cn.
\IEEEcompsocthanksitem 
C. Long works at Nanyang Technological University. E-mail: c.long@ntu.edu.sg.
\IEEEcompsocthanksitem 
L. Su is from Tsinghua University. E-mail: sulc21@mails.tsinghua.edu.cn.
\IEEEcompsocthanksitem 
T. Wu works at Westlake University. E-mail: wutailin@westlake.edu.cn.
\IEEEcompsocthanksitem 
H. Yin works at the University of Queensland. E-mail: h.yin1@uq.edu.au.
\IEEEcompsocthanksitem
J. Chang and Q. Tian work at Huawei Cloud \& AI. E-mail: \{jianlong.chang, tian.qi1\}@huawei.com.
\IEEEcompsocthanksitem
C. Jensen works at Aalborg University. E-mail: csj@cs.aau.dk.
}
}
\begin{document}

\maketitle

\begin{abstract}
This paper focuses on the integration of \emph{generative techniques} into spatial-temporal data mining, considering the significant growth and diverse nature of spatial-temporal data. With the advancements in RNNs, CNNs, and other \emph{non-generative techniques}, researchers have explored their application in capturing temporal and spatial dependencies within spatial-temporal data. However, the emergence of \emph{generative techniques} such as LLMs, SSL, Seq2Seq and diffusion models has opened up new possibilities for enhancing spatial-temporal data mining further. The paper provides a comprehensive analysis of generative technique-based spatial-temporal methods and introduces a standardized framework specifically designed for the spatial-temporal data mining pipeline. By offering a detailed review and a novel taxonomy of spatial-temporal methodology utilizing \emph{generative techniques}, the paper enables a deeper understanding of the various techniques employed in this field. Furthermore, the paper highlights promising future research directions, urging researchers to delve deeper into spatial-temporal data mining. It emphasizes the need to explore untapped opportunities and push the boundaries of knowledge to unlock new insights and improve the effectiveness and efficiency of spatial-temporal data mining. By integrating \emph{generative techniques} and providing a standardized framework, the paper contributes to advancing the field and encourages researchers to explore the vast potential of \emph{generative techniques} in spatial-temporal data mining.

\end{abstract}

\input{intro}

\input{relate}

\input{data}

\input{preli}

\input{solution2}

\input{taxonomy}
\input{eval}

\input{future}

\input{conclusion}

\bibliographystyle{IEEEtran}
\bibliography{sample-base}

\end{document}

%% file: intro.tex
\section{Introduction}
\label{sec:intro}
\IEEEPARstart{W}{ith} the remarkable advancements in GPS technology and mobile devices, the volume of spatial-temporal data has experienced a substantial expansion. This encompassing growth encompasses diverse data types, including human trajectory data, traffic trajectory data, crime data, climate data, event data, and so on. The data mining of spatial-temporal data holds great significance in numerous domains, such as urban management~\cite{engin2020data,button2002city}, where it facilitates effective decision-making processes. Furthermore, it plays a crucial role in optimizing taxi dispatching operations~\cite{liu2020context,lee2004taxi}, leading to improved transportation efficiency. Additionally, the examination of spatial-temporal data has profound implications for human health~\cite{kampa2008human,abrahams2002soils}, enabling the identification of correlations that contribute to enhanced well-being. Notably, it is also pivotal in the realm of weather forecasting, where accurate predictions can dramatically mitigate the impact of natural disasters~\cite{bi2022pangu}.

\begin{figure}
  \centering
  \includegraphics[width=1.02\linewidth]{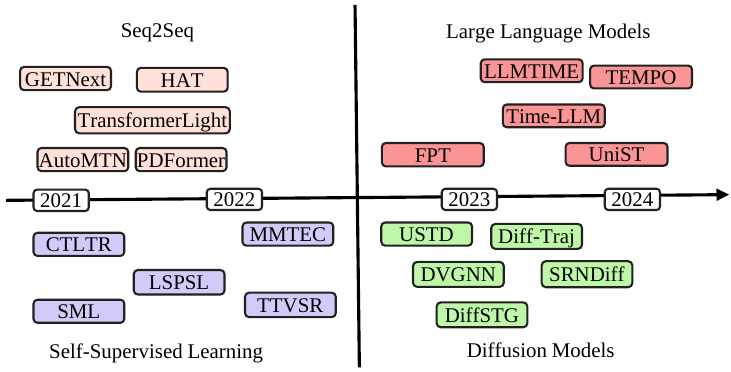}
  \caption{Examples of existing studies via \textit{generative techniques.}
  }   \label{fig:intro}
\end{figure}

In recent years, the success of Recurrent Neural Networks (RNNs)~\cite{schuster1997bidirectional} and Convolutional Neural Networks (CNNs)~\cite{krizhevsky2012imagenet} in capturing temporal and spatial dependencies within spatial-temporal data has been remarkable. This achievement has motivated researchers to further explore and embrace the use of RNNs, CNNs, and other \emph{non-generative techniques} to analyze spatial-temporal data. These efforts have resulted in significant advancements in spatial-temporal tasks, such as traffic prediction~\cite{DCRNN,STGCN} and anomalous trajectory detection~\cite{zhang2023online}. More recently, with the emergence of successful \emph{generative techniques} like LLMs and Diffusion Models (DMs) in the domains of computer vision (CV)~\cite{zhou2022learning,yu2023visual} and natural language processing (NLP)~\cite{duan2020study}, researchers have started exploring whether \emph{generative techniques} can further enhance the performance of spatial-temporal data mining~\cite{liu2024spatial}. This new direction of research has brought fresh insights into spatial-temporal data mining, including zero-shot forecasting and strong generalization capabilities across diverse tasks.

The remarkable accomplishments achieved by \emph{generative techniques} such as LLMs, DMs, and Self-Supervised Learning (SSL) in the domains of CV and NLP have not only inspired researchers but have also had a profound impact. Consequently, there has been a notable surge in the exploration and application of \emph{generative techniques} within the realm of spatial-temporal data mining. Promisingly, recent studies~\cite{liu2024spatial,zhang2023automated} have provided compelling evidence of the advantageous effects of integrating \emph{generative techniques} into spatial-temporal data mining methodologies, leading to notable enhancements in performance. These findings have ignited a heightened interest among researchers, catalyzing a vigorous pursuit of investigating the inherent potential, advantages, and diverse applications of \emph{generative techniques} in this field. As a result, the research landscape has witnessed a prolific emergence of a substantial body of spatial-temporal data mining endeavors rooted in the utilization of \emph{generative techniques} in recent years.

Despite that
numerous studies have focused on spatial-temporal data mining using \emph{generative techniques} (with some representative ones shown in Figure~\ref{fig:intro}), existing surveys~\cite{atluri2018spatio} in this domain lack an extensive analysis and standard framework specifically dedicated to spatial-temporal methodology employing \emph{generative techniques}.
Therefore, we aim to
provide an extensive analysis of methods based on \emph{generative techniques} for spatial-temporal data mining. This analysis not only offers a standardized framework for the spatial-temporal data mining pipeline but also highlights recent pioneering studies utilizing \emph{generative techniques}.
Table~\ref{tab:survey} summarizes the major differences between our paper and existing surveys in this field.

In this paper, we provide an extensive analysis on spatial-temporal methods based on \emph{generative techniques}. Inspired by~\cite{wang2020deep}, we also provide a standardized framework for the spatial-temporal data mining pipeline with \emph{generative techniques}. To sum up, our contributions are presented as follows:
\begin{itemize}[leftmargin=*]

\item \textbf{Extensive and Contemporary Review of Spatial-Temporal Methodology via \emph{Generative Techniques} with Innovative Taxonomy}. Our offering includes a comprehensive and up-to-date review that explores the complexities of spatial-temporal methodology. We meticulously explore a diverse array of generative techniques employed in the context of spatial-temporal data mining. Moreover, we introduce a novel and thoughtfully crafted taxonomy that encompasses the multifaceted nature of spatial-temporal data and encompasses an extensive range of \emph{generative techniques}-based spatial-temporal studies.

\begin{table}[!t]
\setlength{\tabcolsep}{2pt}
\centering
\caption{In comparing our survey to related surveys in the field, several key distinctions and contributions emerge. 
}
\small
\setlength{\tabcolsep}{1.3mm}{
\begin{tabular}{lcccccccc}
\toprule
Survey         & \cite{geetha2008survey} & \cite{qunyong2016survey} & \cite{wang2020deep} & \cite{atluri2018spatio} & \cite{mahmood2019spatio} &  \cite{hamdi2022spatiotemporal}&\cite{sharma2022spatiotemporal}& \textbf{Ours} \\ \hline
Taxonomy       &No   &Yes   &No   &No   &Yes   &Yes     &No   & Yes \\
Pipeline       &No   &No   &Yes   &No   &No   &No     &Yes & Yes \\
\textit{Seq2Seq} &No   &Yes   &Yes   &Yes   &Yes   &Yes     &Yes & Yes \\
\textit{SSL} &No   &No   &Yes   &No   &No   &No     &No & Yes \\
\textit{DMs} &No   &No   &No   &No   &No   &No     &No & Yes \\
\textit{LLMs} &No   &No   &No   &No   &No   &No     &No & Yes \\
\textit{Year}       &\textit{2008}   &\textit{2016}  &\textit{2020}   &\textit{2018}   &\textit{2019}   &\textit{2022}     &\textit{2022}   & \textit{2024} \\
\bottomrule
\end{tabular}}
\label{tab:survey}
\end{table}

\item \textbf{Innovative Integration of Spatial-Temporal Methodology with Standardized Framework}. 
We provide our innovative offering, which brings together the integration of spatial-temporal methodology with a standardized framework. In addition to offering a comprehensive exploration of spatial-temporal methodology, our solution takes a step further by incorporating a standardized framework. This framework provides a systematic pipeline for selecting suitable \emph{Generative Techniques} in the realm of spatial-temporal data mining. 

\item \textbf{Numerous Promising Avenues for Future Research}. We thoroughly emphasize the potential future directions for spatial-temporal data mining. We strongly encourage researchers to investigate deeper into this promising field, exploring untapped opportunities and pushing the boundaries of knowledge. By doing so, we can unlock new insights and further enhance the effectiveness and efficiency of spatial-temporal data mining.

\end{itemize}

The remainder of this paper is structured as follows:
Sec. \ref{sec:relate} clearly articulate the related work and our motivation.
Sec. \ref{sec:data} explores data perspectives, emphasizing the unique challenges and solutions associated with spatio-temporal data.
Sec.~\ref{sec:preli} offers a detailed overview of LLMs, DMs, SSL and Seq2Seq, elucidating their evolution, underlying theoretical principles, and diverse implementations.
Sec. \ref{sec:solution} presents a structured overview of spatial-temporal mining, setting the
stage for a deeper exploration of application perspectives in Sec. \ref{sec:eval}, which explores
the application of existing methods across various tasks, such as representation learning, forecasting, recommendation, clustering, and so on. Most importantly, in Sec. \ref{sec:future}, we conduct a thorough analysis of the shortcomings of existing methods and identify potential trends, outlining a clear path for future development. Finally, Sec. \ref{sec:conclusoin} concludes the paper with summarizing remarks.

%% file: relate.tex
\section{Related Works on Existing Surveys}
\label{sec:relate}
Surveys in the field of spatial-temporal data analysis can be broadly divided into two categories. The first category focuses on spatial-temporal forecasting with deep learning methods or targeting for GNN-based methods~\cite{jin2023spatio,luo2023stg4traffic}. To be specific, Jin \textit{\textit{et al.}}\cite{jin2023spatio} provides a comprehensive overview of the advancements in Spatial-Temporal Graph Neural Networks (STGNNs) for predictive learning in urban computing. Another recent study by Luo \textit{\textit{et al.}}\cite{luo2023stg4traffic} conducts a systematic review of graph learning strategies and prevalent graph convolution algorithms, specifically in the context of spatial-temporal prediction. The authors of~\cite{luo2023stg4traffic} also thoroughly analyze the strengths and limitations of previously proposed models for spatial-temporal graph networks. Furthermore, Luo \textit{\textit{et al.}}~\cite{luo2023stg4traffic} establish a benchmark using the PyTorch deep learning framework to facilitate performance evaluation.

The second category of research in spatial-temporal data mining comprises surveys and comprehensive overviews~\cite{geetha2008survey,qunyong2016survey,atluri2018spatio,wang2020deep,mahmood2019spatio,hamdi2022spatiotemporal,sharma2022spatiotemporal} that complement the first research line mentioned earlier. Geetha \textit{et al.}~\cite{geetha2008survey} conducts a comprehensive analysis of spatial-temporal data and their attributes up until 2008, but does not provide a taxonomy and pipeline. Similarly, Qunyong \textit{et al.}~\cite{qunyong2016survey} aims to provide an overview of diverse spatial-temporal data mining techniques, offering a taxonomy of existing methods until 2016. Atluri \textit{et al.}~\cite{atluri2018spatio} surveys different methods employed for spatial-temporal data mining until 2018. While Hamdi \textit{et al.}~\cite{hamdi2022spatiotemporal} presents a study highlighting challenges and research problems in spatial-temporal data mining, none of these surveys included up-to-date \textit{generative techniques}. Moreover, none of them provide both a taxonomy and a standardized pipeline for spatial-temporal mining.

Another research line~\cite{alam2022survey,pant2018survey,abraham1999survey,jitkajornwanich2020survey} focuses on spatial-temporal databases, which is different from the former two research categories on spatial-temporal prediction and spatial-temporal data mining. Similar to the first research line, this work targets providing a survey on spatial-temporal data mining problems and corresponding methods. And different from the first research line, we mainly focus on up-to-date generative methods, including LLMs, self-supervised learning, and diffusion models on spatial-temporal data.

\textbf{Motivation of This Survey}. In contrast to the previous surveys of the spatial-temporal data mining presented, which is shown in Table~\ref{tab:survey}, our emphasis is on the most recent advancements in \textit{generative techniques} for spatial-temporal data. Our review encompasses a comprehensive and up-to-date analysis of various approaches, including LLMs, SSL, DMs, and Seq2Seq, applied specifically to spatial-temporal data mining. We introduce an innovative taxonomy to categorize these methods effectively. By exploring the potential of these cutting-edge \textit{generative techniques}, we aim to drive advancements in the field of spatial-temporal data mining. Additionally, we provide a standardized framework with an innovative taxonomy that combines different spatial-temporal methods.

%% file: data.tex
\section{Data}
\label{sec:data}
Spatial-temporal data encompasses various types, each distinguished by unique data collection methods in real-world applications. The wide range of application scenarios and spatial-temporal data types results in different categories of data mining tasks and problem formulations. Moreover, diverse spatial-temporal data mining methods exhibit varying preferences for specific types of spatial-temporal data and have distinct requirements for input data formats. For example, temporal convolutional neural networks excel at capturing temporal dependencies, while graph convolutional neural networks are good at capturing spatial correlations. In this section, we begin by introducing the properties of spatial-temporal data. Subsequently, we follow the approach presented in the study by ~\cite{atluri2018spatio}. Atluri \textit{et al.}~\cite{atluri2018spatio} classify spatial-temporal data into four distinct groups: event data, trajectory data, point data, and raster data. Furthermore, we provide illustrative examples of spatial-temporal data instances and showcase different approaches for representing such data.

\subsection{Data Properties} 
Spatial-temporal data poses two significant challenges, which simultaneously serve as issues and opportunities for spatial-temporal data mining algorithms. These challenges encompass correlations and skewed distributions.

\noindent \textbf{Correlations:} Spatial-temporal correlations refer to the inter-dependencies and relationships that exist between different spatial and temporal aspects within a dataset due to collection mechanism of spatial-temporal data. In real-world applications, these correlations can give rise to several issues and complexities. Firstly, \underline{prediction accuracy}, spatial-temporal correlations can introduce intricate patterns and dependencies within the data. Failing to capture and model these correlations accurately can result in reduced prediction accuracy, hindering the effectiveness of predictive models in various domains such as traffic forecasting, weather prediction, and disease outbreak analysis. Secondly, \underline{data preprocessing and fusion}, spatial-temporal correlations often necessitate the integration and fusion of data from multiple sources and modalities. Combining heterogeneous data types and aligning them in a spatial-temporal context can be challenging, requiring careful preprocessing steps and fusion techniques to ensure the accuracy and consistency of the data.

\noindent \textbf{Heterogeneity:} Spatial-temporal heterogeneity refers to the inherent variability and diversity of spatial and temporal patterns within a dataset. It reflects the fact that different regions and time periods exhibit distinct characteristics, trends, and relationships in spatial-temporal data. This heterogeneity can give rise to several issues in data analysis and pose challenges for spatial-temporal data mining algorithms. Some of the key issues associated with spatial-temporal heterogeneity include:
 Firstly, \underline{generalization challenges}, spatial-temporal heterogeneity makes it challenging to develop generalized models and algorithms that can effectively capture and represent the diverse patterns and relationships across different regions and time periods. Models trained on one region or time period may not generalize well to other regions or time periods due to the spatial-temporal variations. Secondly, \underline{biases and incompleteness}, spatial-temporal heterogeneity can lead to biases and incompleteness in data collection and representation. Data may be unevenly distributed across regions and time periods, resulting in imbalanced datasets. This can introduce biases and skew the analysis and results of spatial-temporal data mining algorithms.

\subsection{Data Types} Spatial-temporal data combines spatial and temporal components, providing insights into phenomena across space and time. It is prevalent in domains like environmental monitoring, transportation analysis, epidemiology, social sciences, and urban planning. This data is represented using geographic coordinates and timestamps, enabling analysis of patterns, trends, and relationships. It can be categorized into event data (discrete events), trajectory data (sequences of sampled points), point data (continuous spatial-temporal field measurements), and raster data (observations in fixed cells). In the following sections, we will introduce each of these data types in detail.

\begin{figure}
\centering
\includegraphics[width=0.9\linewidth]{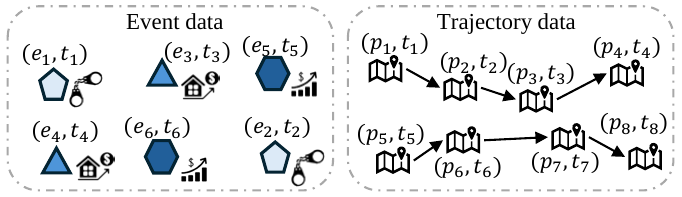}
\caption{\emph{Example of Event data and trajectory data}}
\label{fig:data_event}
\end{figure}

\noindent \textbf{Event data:} Event data can be characterized as specific occurrences taking place at particular locations and times. Examples of event data include crime data or voting data, where each event is associated with a spatial-temporal location. Apart from the spatial-temporal aspects, event data can also include additional variables known as marked variables. These marked variables provide supplementary information about the event, such as the type of crime in crime data or the political party a person has voted for in voting data. Typically, event data is represented using the Euclidean coordinate system, which measures distances in a straight line. However, in cases where events occur within a road network, such as accidents, the distance between two events is determined by the shortest path along road segments rather than the Euclidean distance. Figure~\ref{fig:data_event} demonstrates a example of three different types of event data, namely crime event, house price event and sales event.

\noindent \textbf{Trajectory data:} Trajectories refer to the spatial paths followed by objects over a period of time. Examples of trajectory data include flight data and taxi data. Typically, trajectory data is gathered by attaching a sensor onto the moving object, which records GPS positions at various time intervals. The accuracy of the trajectory is enhanced when the time intervals between these recorded positions are minimized. We provide an example in Figure~\ref{fig:data_event}. Figure~\ref{fig:data_event} illustrates two trajectories, which are composed of a sequence of points, denoted as $\left\{(p_1, t_1), (p_2, t_2), ..., (p_8, t_8)\right\}$. In this context, $p_i$ represents the location, indicated by latitude and longitude coordinates. Trajectory data has become increasingly prevalent with the advent of mobile applications equipped with GPS devices.

\begin{figure}
\centering
\includegraphics[width=0.9\linewidth]{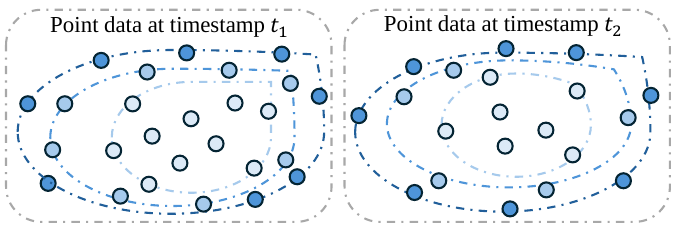}
\caption{\emph{Example of Point data at different timestamps}}
\label{fig:data_point}
\end{figure}

\noindent \textbf{Point data:} Point data refers to data obtained from a collection of mobile reference points. For example, this could include data gathered by weather balloons hovering in the atmosphere or sensors measuring the temperature of a water body's surface. Let's consider a scenario where temperature measurements are recorded at various weather stations across a region. Each weather station represents a specific point location, and temperature readings are taken at regular intervals. An example on point data is shown in Figure~\ref{fig:data_point}. Different dashed lines represent the distributions of the continuous spatial-temporal field. we can observe the spatial arrangement of a continuous spatial-temporal field at two distinct timestamps, which are measured at specific point locations represented by blue points on each timestamp. These discrete blue points provide valuable observations that can be utilized to reconstruct the spatial-temporal field at any desired location and timestamp.

\begin{figure}
\centering
\includegraphics[width=0.9\linewidth]{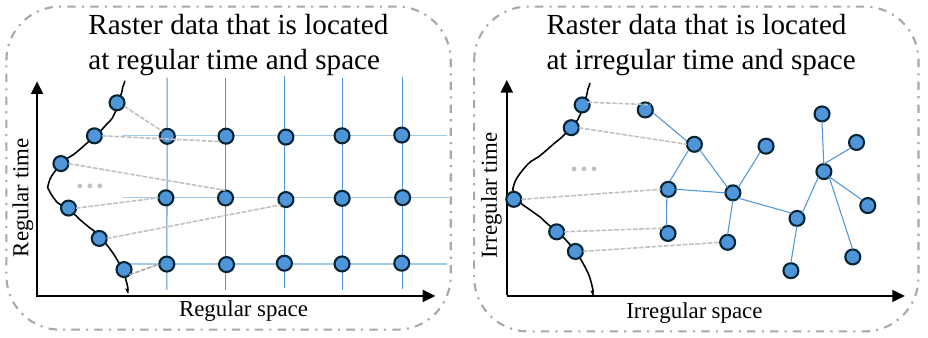}
\caption{\emph{Example of Raster data at regular time and place and irregular time and space}}
\label{fig:data_raster}
\end{figure}

\noindent \textbf{Raster data:} In raster data, reference points remain static, unlike mobile reference points in point reference data. These fixed locations can be regularly or irregularly distributed in space, as seen in pixels of images or ground-based sensor systems. Observations in raster data are recorded at fixed time intervals, regularly or irregularly spaced. Satellite imagery is a common raster data type, capturing images where each pixel corresponds to a cell in the raster grid, containing information about spectral characteristics. Digital Elevation Models (DEMs) represent elevation or topography, with each grid cell holding height information. Raster data is also used for climate variables, representing temperature, precipitation, and wind speed, where each cell represents a location with corresponding climate attributes. Land cover or land use information is another application of raster datasets, with each cell representing a specific location and displaying land cover classes like forest, urban area, water body, or agricultural land. We provide an example in Figure~\ref{fig:data_raster}. The left figure shows the location set and timestamp set of the raster data, which are located regularly. The right figure represent that the raster data are located regularly in the set of locations and timestamps. The grey dashed lines connect the same point in regular space with regular time.

\subsection{Data Instances}
Data mining algorithms operate on data instances, which are specific examples or observations in a dataset used for analysis and modeling. Instances represent individual rows or records with attribute values. In spatial-temporal data mining, different approaches define instances for specific data types, leading to distinct formulations. We explore prevalent categories of spatial-temporal (ST) instances: points, trajectories, time series, spatial maps, and spatial-temporal rasters. 
We provide the mapping relations between data type and data instance in Figure~\ref{fig:data_map}. These instances reveal diverse applications of spatial-temporal data mining. Spatial-temporal events can be represented as point instances, while trajectory data can be represented as point collections or as time series of spatial identifiers. Spatial-temporal reference data uses point instances as reference points. For spatial-temporal raster data, instances can be constructed as time series at a location, spatial maps at a specific time, or the entire raster as a single instance. The choice of instance construction depends on the research question and applicable spatial-temporal data mining methods.

%% file: preli.tex
\begin{figure}
\centering
\includegraphics[width=0.9\linewidth]{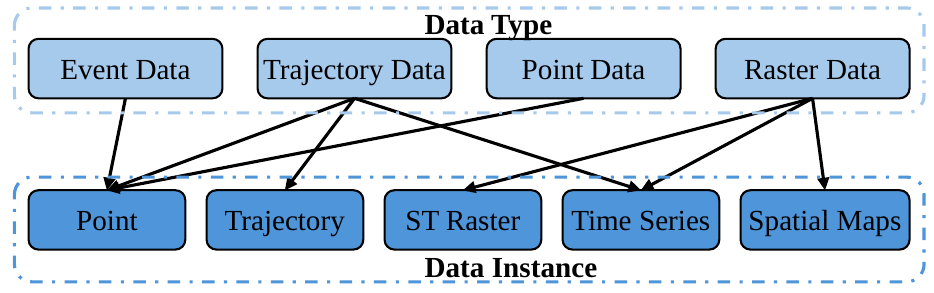}
\caption{\emph{Mapping between data type and data instance}}
\label{fig:data_map}
\end{figure}

\section{Preliminary on \textit{Generative Techinques}}
\label{sec:preli}
In this section, we aim to provide illustration on \textit{generative techniques}, including LLMs, DMs, and Seq2Seq.
\subsection{Large Language Models (LLMs)} 
LLMs are trained using extensive text data and have shown impressive capabilities in CV~\cite{zhou2022learning,zhu2023minigpt,zhou2022conditional,zhao2023learning,wang2024parameter} and NLP~\cite{duan2020study,mathew2020review,gu2021domain,jiang2024hum,sun2024triforce}. They rely on the transformer architecture, which includes encoder and decoder modules with self-attention mechanisms~\cite{han2021transformer}. LLMs can be classified into three categories based on their architectural structure~\cite{pan2023unifying}: only-encoder, encoder-decoder, and only-decoder models. (1) Only-encoder LLMs, like BERT~\cite{devlin2018bert,liu2019roberta,lan2019albert}, use the encoder module to understand word associations and encode sentences. A secondary prediction head is added for downstream tasks. These models excel in tasks that require a comprehensive understanding of the entire sentence~\cite{zhang2022survey}, such as named entity recognition~\cite{settles2004biomedical} and text categorization~\cite{yao2019kg,hakala2019biomedical}. (2) Encoder-decoder LLMs employ both encoder and decoder modules. The encoder encodes input sentences into a hidden space, while the decoder generates the desired output text~\cite{raffel2020exploring,zeng2022glm}. Training techniques for these models offer more flexibility~\cite{zoph2022st,xue2020mt5} in generating diverse outputs. (3) Only-decoder LLMs solely rely on the decoder module to generate output by predicting the next token in each sentence. Large-scale only-decoder LLMs perform effectively even with limited samples or a few instructions on downstream tasks, which means that further fine-tuning is not needed~\cite{brown2020language}. While modern LLMs like Chat-GPT~\cite{ouyang2022training} and GPT-4 commonly adopt a only-decoder architecture, they are often proprietary and not easily accessible to academic researchers, posing challenges for further exploration~\cite{brown2020language}. However, recently, open-source only-decoder LLMs such as Alpaca and Vicuna have been made available. These models have been modified based on existing literature and demonstrate comparable performance to Chat-GPT and GPT-4, providing researchers with accessible alternatives for experimentation and development.

\subsection{Diffusion Models (DMs)} 
\begin{figure*}
  \centering
  \includegraphics[width=\linewidth]{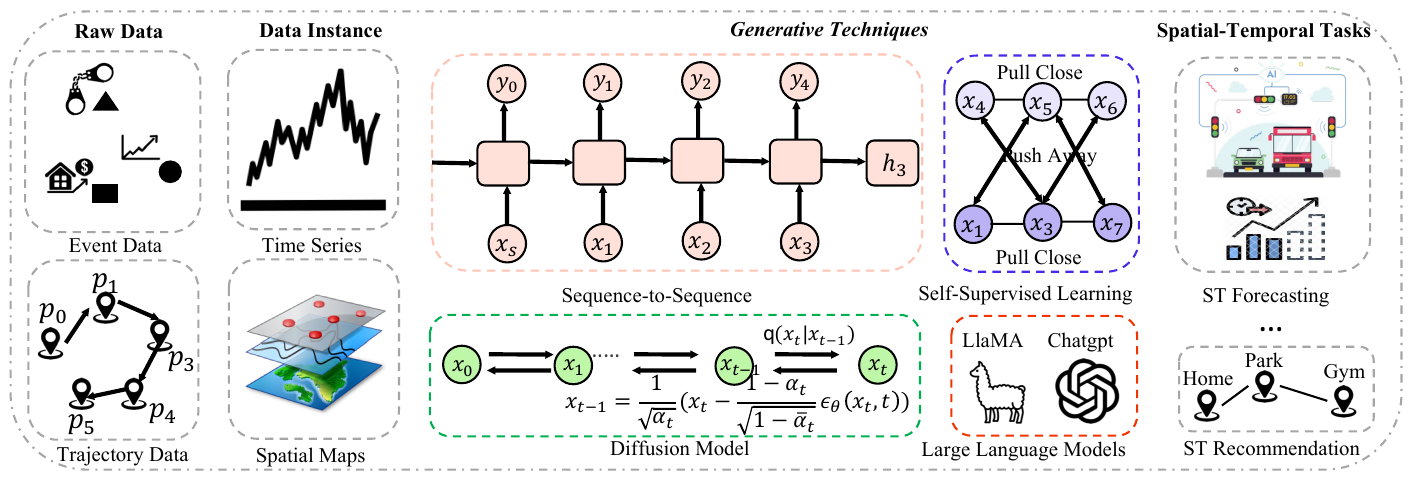}
  \caption{Framework of Spatial-Temporal Data Mining via \textit{generative techniques.}
  }   
  \label{fig:fram}
\end{figure*}

The Denoising Diffusion Probabilistic Model (DDPM) \cite{ho2020denoising} consists of two fundamental processes: the forward process, also known as the diffusion process, and the reverse process, which is shown in Figure~\ref{fig:fram}. Let's start by focusing on describing the forward process. In a diffusion model, the forward process approximates the posterior distribution $q(x_{1:T}|x_0)$, which represents the sequence of latent variables $x_{1:T}$ given an initial value $x_0$. This approximation is achieved by applying a Markov chain iteratively, gradually introducing Gaussian noise over time.

The forward process is represented as follows:
\begin{align}
\label{eq:posterior}
q(x_t|x_{t-1}) = \mathcal{N}\left(x_t; \mu_t(x_{t-1}), \beta_t\mathbf{I}\right)
\end{align}
Here, $x_t$ denotes the latent variable at time step $t$, and $x_{t-1}$ is the variable at the previous time step. The distribution $q(x_t|x_{t-1})$ is modeled as a Gaussian distribution with a mean $\mu_t(x_{t-1})$ and a variance parameter $\beta_t\mathbf{I}$, where $\beta_t$ represents the variance at time step $t$, and $\mathbf{I}$ denotes the identity matrix. The mean $\mu_t(x_{t-1})$ can depend on the previous latent variable $x_{t-1}$ and is typically modeled using neural networks or other parameterized functions.

By sequentially applying the distribution $q(x_t|x_{t-1})$ for each time step, starting from the initial value $x_0$, we obtain an approximation of the posterior distribution $q(x_{1:T}|x_0)$. This approximation captures the temporal evolution of the latent variables by computing $q(x_{1:T}|x_0) := \prod_{t=1}^{T}q(x_{t}|x_{t-1})$. To explain the reverse process, let's consider a diffusion model with $T$ time steps. The objective is to generate a sample from the initial distribution $p(x_0)$ given an observed data point $x_{T}$ at the final time step. The reverse process in a diffusion model can be described as follows:
1. Initialization: Set $x_{T}$ as the observed data point.
2. Iterative Sampling: Starting from $t = T-1$ and moving backwards until $t = 0$, sample $x_t$ from the distribution $p(x_t|x_{t+1})$, which represents the reverse diffusion process.

In the reverse process, the distribution $p(x_t|x_{t+1})$ is typically modeled as a Gaussian distribution, similar to the forward process. However, the mean and variance parameters are adjusted to account for the reverse direction. The specific form of $p(x_t|x_{t+1})$ is defined as follows:
\begin{align}
\label{eq:forward}
p_{\theta}(x_{0:T}) &:= p(x_T) \prod_{t=1}^{T}p_{\theta}(x_{t-1}|x_t);\nonumber\\
p(x_t|x_{t+1}) &:= \mathcal{N}(x_{t}; \mu_{\theta}(x_{t+1}, t+1), \sum_{\theta}(x_{t+1}, t+1))
\end{align}

By iteratively sampling from the reverse process, we can generate a sequence of latent variables $x_{0:T}$ that follows the reverse diffusion process. This sequence represents a sample from the initial distribution $p(x_0)$. The reverse process plays a crucial role in training the diffusion model. During training, the model learns to approximate the reverse process by minimizing the difference between the generated samples and the observed data points. This training procedure ensures that the model captures the underlying data distribution and can generate realistic samples. The optimization objective of the diffusion model is achieved through the negative log likelihood, which can be expressed as:
\begin{align}
\label{eq:ddpm_loss}
&\mathbb{E}[-\log p_{\theta}(x_0)] \leq \mathbb{E}_{q} [- \log \frac{p_{\theta}(x_{0:T})}{q(x_{1:T}|x_0)}] \nonumber\\
& = \mathbb{E}_{q}[-\log p(x_T) - \sum_{t \geq 1} \log \frac{p_{\theta}(x_{t-1}|x_t)}{q(x_t|x_{t-1})}]
\end{align}

\subsection{Self-Supervised Learning (SSL)} SSL is a versatile learning framework that operates by leveraging surrogate tasks, which can be formulated using unsupervised data alone. These pretext tasks are carefully designed to necessitate the learning of valuable representations for the data. Self-supervised learning techniques find applications across a wide spectrum of computer vision domains and topics~\cite{jang2018grasp2vec,sermanet2018time,ebert2018robustness,owens2018audio,sayed2019cross}. They enable advancements in areas such as image recognition, object detection, image segmentation, and many others, by capitalizing on the abundance of unlabeled data available. Through self-supervised learning, models can acquire meaningful representations, learn useful features, and generalize well to various downstream tasks without the need for explicit supervision.

SSL is divided into two categories. The first category is self-supervised contrastive learning~\cite{liu2021self}. The main idea behind contrastive learning is to ensure that representations align with each other when subjected to appropriate transformations. This idea has recently gained significant attention in the field of visual representation learning, leading to a surge of research efforts~\cite{becker1992self,wu2018unsupervised,ye2019unsupervised,ji2019invariant,chen2020simple}. Similarly, in the context of graph data, traditional methods aimed at reconstructing vertex adjacency information~\cite{kipf2016variational,hamilton2017inductive} can be viewed as a form of "local contrast." However, these methods tend to overly emphasize proximity details while neglecting structural information~\cite{ribeiro2017struc2vec}. Inspired by~\cite{belghazi2018mine,hjelm2018learning}, recent proposals~\cite{ribeiro2017struc2vec,sun2019infograph,peng2020self} suggest the application of contrastive learning to capture structural information by contrasting local and global representations. Nevertheless, the exploration of contrastive learning in graphs has not yet considered the enforcement of perturbation invariance, as has been done in previous works~\cite{ji2019invariant,chen2020simple}. Another category is generative self-supervised learning~\cite{liu2021self}. Generative self-supervised learning focuses on training models to generate high-quality and meaningful data representations in an unsupervised manner. Unlike discriminative self-supervised learning, which relies on pretext tasks, generative self-supervised learning leverages the power of generative models to learn representations that capture the underlying data distribution.

Another category is generative, self-supervised learning. By modeling the data generation process, generative models like variational autoencoders (VAEs) can learn to map useful information into the latent space. These learned representations can then be utilized for downstream tasks such as image generation, data augmentation, and semantic understanding. Several notable works have contributed to the development of generative SSL. For example, InfoGAN~\cite{chen2016infogan} introduces an information-theoretic extension to GANs, enabling disentangled representation learning. Similarly, $\beta$-VAE~\cite{higgins2017beta} introduces a regularization term to encourage disentanglement in VAEs. Adversarial autoencoders (AAEs)~\cite{makhzani2015adversarial} combines the strengths of VAEs and GANs to learn generative representations with improved diversity and quality. Other works have explored the application of generative self-supervised learning in specific domains. For instance, in CV, DeepCluster~\cite{caron2018deep} proposes clustering-based methods for unsupervised feature learning.

\subsection{Sequence-to-Sequence (Seq2Seq)} 
Sequence-to-Sequence (Seq2Seq)~\cite{sutskever2014sequence} models have garnered significant attention in the domain of NLP due to their remarkable ability to handle tasks involving sequential data. We provide a brief structure of Seq2Seq in Figure~\ref{fig:fram}. Seq2Seq models provide a powerful framework for transforming an input sequence into an output sequence, making them particularly suitable for applications like machine translation, text summarizing, and speech recognition.

The Seq2Seq models are built upon the encoder-decoder architecture, which has been widely studied and adopted in various NLP tasks. The input sequence is fed into the encoder and some essential context information is captured. This learned hidden representation is then passed to the decoder, which generates the output sequence step by step, taking into account the context provided by the encoder. The decoder can employ autoregressive techniques, generating one token at a time conditioned on the previously generated tokens, or utilize advanced search strategies such as beam search to explore multiple possibilities and select the most likely sequence.

Combination of the attention mechanism makes a large improvement to the success of Seq2Seq models on many domains. Attention allows the decoder to pay attention to different parts of the input sequence dynamically while generating each output part. This has a very positive effect on capturing long-range dependencies. The attention mechanism has proven to be a crucial component in achieving state-of-the-art results in Seq2Seq models across a range of tasks. Seq2Seq models have demonstrated impressive performance in machine translation. For instance, the attention-based neural machine translation model proposed by Bahdanau \textit{et al.}~\cite{bahdanau2014neural} significantly improves translation quality. Seq2Seq models have also been successfully applied to text summarization, as showcased in the work of Nallapati \textit{et al.}~\cite{nallapati2016abstractive}, where they generates concise and informative summaries. Furthermore, in speech recognition, Seq2Seq models have shown promise in converting spoken language to written text, as demonstrated by Chan \textit{et al.}~\cite{weiss2017sequence}.

%% file: solution2.tex
\section{Framework}
\label{sec:solution}

In this section, we present an approach to tackle spatial-temporal data mining challenges using \textit{generative techniques}. We begin by discussing the preprocessing of the data, followed by an adaptation introduction to \textit{generative techniques}. Furthermore, we dedicate a subsection to address specific spatial-temporal data mining problems. To provide a structured overview, we offer a framework outlining the pipeline, which is visualized in Figure~\ref{fig:fram}.

Figure~\ref{fig:fram} illustrates a general pipeline for utilizing \textit{generative techniques} in spatial-temporal data mining. The pipeline involves processing raw spatial-temporal data collected from diverse location sensors, including event data, trajectory data, point reference data, and raster data. Initially, data instances are created to store the spatial-temporal data, which we have previously discussed. These instances can take the form of points, time series, spatial maps, trajectories, or spatial-temporal rasters. To employ \textit{generative techniques} for different mining tasks, the spatial-temporal data instances need to be transformed into specific data formats. The spatial-temporal data instances can be represented as sequence data, matrices, tensors, or graphs, depending on the data representation chosen. Finally, the selected \textit{generative techniques} are applied to address a variety of spatial-temporal data mining tasks, such as prediction, classification, and representation learning, among others. These models leverage the unique capabilities of \textit{generative techniques} to extract valuable insights from spatial-temporal data and tackle complex spatial-temporal data mining challenges.

\subsection{Spatial-Temporal Data Preprocessing}
Spatial-temporal data preprocessing is crucial for analyzing and extracting insights from ST datasets. It involves cleaning, transforming, and organizing raw data for subsequent analysis. Objectives include enhancing data quality, handling missing values/outliers, integrating multiple sources, and creating effective representations. Cleaning ensures dataset reliability and accuracy. Integration merges datasets for comprehensive analysis. Handling missing values/outliers maintains data integrity. Transforming data into suitable representations (points, trajectories, time series, spatial maps, or rasters) is essential for effective data mining.

\subsection{The Framework of Spatial-Temporal Data Mining }
We provide the framework shown in Figure~\ref{fig:fram} to present a comprehensive depiction of the general pipeline utilized for harnessing the potential of advanced generative techniques in spatial-temporal data mining. The pipeline entails the processing of raw spatial-temporal data obtained from diverse location sensors, which encompass various data types such as event data, trajectory data, point reference data, and raster data. Initially, specific data instances are created to store the spatial-temporal data, as discussed earlier. These instances can take various forms, including points, time series, spatial maps, trajectories, or spatial-temporal rasters, depending on the nature of the data. To effectively employ generative techniques for different mining tasks, the spatial-temporal data instances must be transformed into specific data formats that align with the chosen generative approach. Depending on the selected data representation, the spatial-temporal data instances can be represented as sequences, matrices, tensors, or graphs. Finally, the chosen generative techniques are applied to address a wide range of spatial-temporal data mining tasks, encompassing prediction, classification, representation learning, and more. These models leverage the unique capabilities of generative techniques to extract valuable insights from spatial-temporal data and overcome the challenges inherent in complex spatial-temporal data mining endeavors.

%% file: taxonomy.tex
\tikzstyle{my-box}=[
    rectangle,
    draw=hidden-draw,
    rounded corners,
    align=left,
    text opacity=1,
    minimum height=1.5em,
    minimum width=5em,
    inner sep=2pt,
    fill opacity=.8,
    line width=0.8pt,
]

\tikzstyle{leaf-head}=[my-box, minimum height=1.5em,
    draw=yellow!80, 
    fill=yellow!35,  
    text=black, font=\normalsize,
    inner xsep=2pt,
    inner ysep=4pt,
    line width=0.8pt,
]

\tikzstyle{leaf-task}=[my-box, minimum height=2.5em,
    draw=red!80, 
    fill=red!25,  
    text=black, font=\normalsize,
    inner xsep=2pt,
    inner ysep=4pt,
    line width=0.8pt,
]

\tikzstyle{leaf-paradigms}=[my-box, minimum height=2.5em,
    draw=blue!70, 
    fill=blue!15,  
    text=black, font=\normalsize,
    inner xsep=2pt,
    inner ysep=4pt,
    line width=0.8pt,
]
\tikzstyle{leaf-others}=[my-box, minimum height=2.5em,
    draw=green!80, 
    fill=green!15,  
    text=black, font=\normalsize,
    inner xsep=2pt,
    inner ysep=4pt,
    line width=0.8pt,
]
\tikzstyle{leaf-other}=[my-box, minimum height=2.5em,
    draw=orange!80, 
    fill=orange!15,  
    text=black, font=\normalsize,
    inner xsep=2pt,
    inner ysep=4pt,
    line width=0.8pt,
]

\tikzstyle{modelnode-task}=[my-box, minimum height=1.5em,
    draw=red!80, 
    text=black, font=\normalsize,
    inner xsep=2pt,
    inner ysep=4pt,
    line width=0.8pt,
]

\tikzstyle{modelnode-paradigms}=[my-box, minimum height=1.5em,
    draw=blue!70, 
    text=black, font=\normalsize,
    inner xsep=2pt,
    inner ysep=4pt,
    line width=0.8pt,
]
\tikzstyle{modelnode-others}=[my-box, minimum height=1.5em,
    draw=green!80, 
    text=black, font=\normalsize,
    inner xsep=2pt,
    inner ysep=4pt,
    line width=0.8pt,
]
\tikzstyle{modelnode-other}=[my-box, minimum height=1.5em,
    draw=orange!80, 
    text=black, font=\normalsize,
    inner xsep=2pt,
    inner ysep=4pt,
    line width=0.8pt,
]
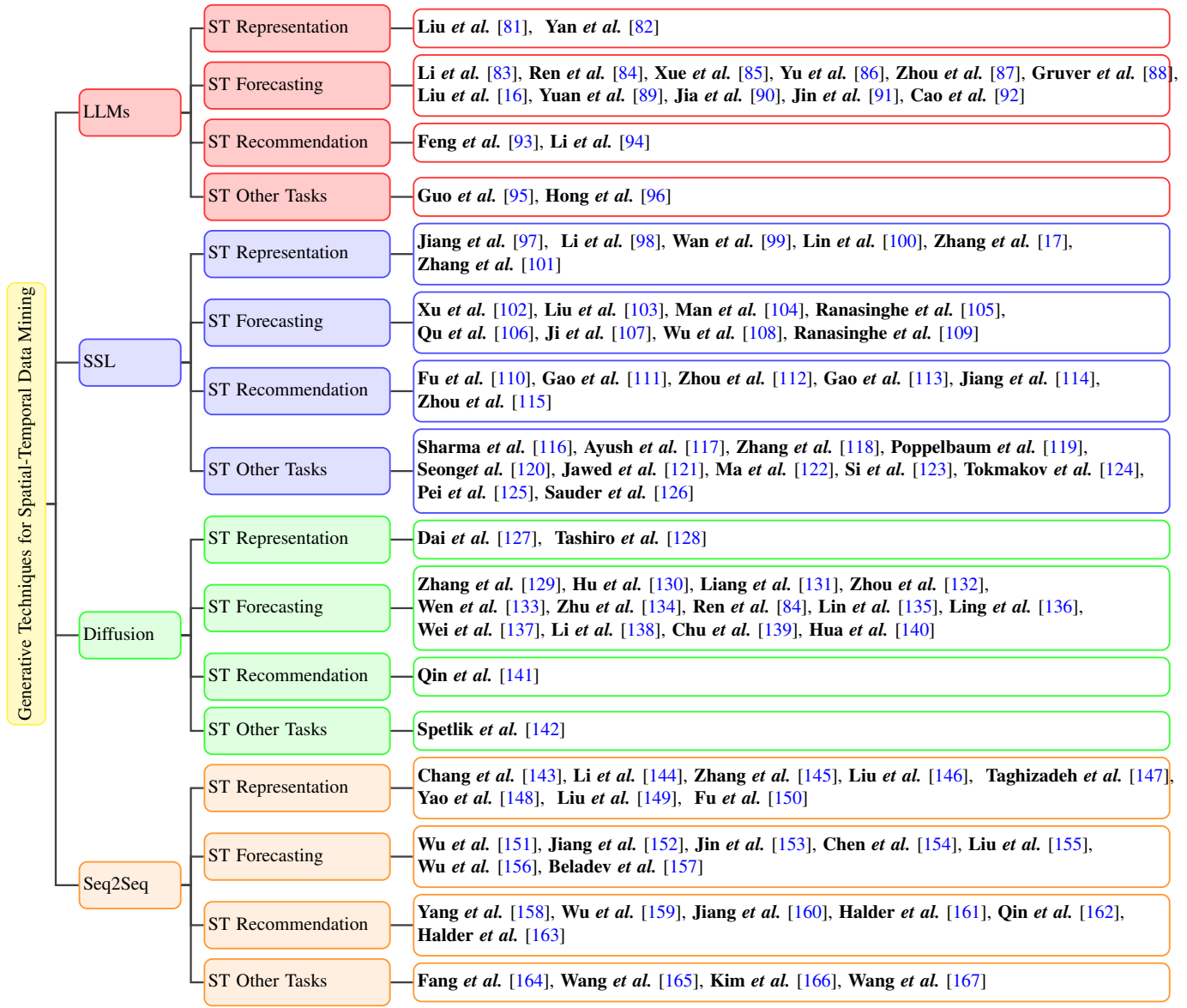
\begin{figure*}[!th]
    \centering
    \resizebox{1\textwidth}{!}{
        \begin{forest}
            forked edges,
            for tree={
                grow=east,
                reversed=true,
                anchor=base west,
                parent anchor=east,
                child anchor=west,
                base=left,
                font=\normalsize,
                rectangle,
                draw=hidden-draw,
                rounded corners,
                align=left,
                minimum width=1em,
                edge+={darkgray, line width=1pt},
                s sep=3pt,
                inner xsep=0pt,
                inner ysep=3pt,
                line width=0.8pt,
                ver/.style={rotate=90, child anchor=north, parent anchor=south, anchor=center},
            }, 
            [
                Generative Techniques for Spatial-Temporal Data Mining,leaf-head, ver
                [
                     LLMs, leaf-task,text width=5em
                    [
                        ST Representation, leaf-task, text width=9.5em
                        [\textbf{Liu \textit{et al.}}~\cite{liu2022cstrm}{, } \textbf{Yan \textit{et al.}}~\cite{yan2023urban}, modelnode-task, text width=40em]
                    ]
                    [
                        ST Forecasting, leaf-task, text width=9.5em
                         [\textbf{Li \textit{et al.}}~\cite{li2022mining}{, }\textbf{Ren \textit{et al.}}~\cite{ren2024tpllm}{, }\textbf{Xue \textit{et al.}}~\cite{xue2023promptcast}{, }\textbf{Yu \textit{et al.}}~\cite{yu2023temporal}{, }\textbf{Zhou \textit{et al.}}~\cite{zhou2024one}{, }\textbf{Gruver \textit{et al.}}~\cite{gruver2024large}{, }\\\textbf{Liu \textit{et al.}}~\cite{liu2024spatial}{, }\textbf{Yuan \textit{et al.}}~\cite{yuan2024unist}{, }\textbf{Jia \textit{et al.}}~\cite{jia2024gpt4mts}{, }\textbf{Jin \textit{et al.}}~\cite{jin2023time}{, }\textbf{Cao \textit{et al.}}~\cite{cao2023tempo}, modelnode-task, text width=40em]
                    ]
                    [
                        ST Recommendation, leaf-task, text width=9.5em
                        [\textbf{Feng \textit{et al.}}~\cite{feng2024move}{, }\textbf{Li \textit{et al.}}~\cite{li2024large}, modelnode-task,text width=40em]
                    ]
                    [
                        ST Other Tasks, leaf-task, text width=9.5em
                        [\textbf{Guo \textit{et al.}}\cite{guo2023point}{, }\textbf{Hong \textit{et al.}}\cite{hong20233d}, modelnode-task,text width=40em]
                    ]
                ]
                [
                    SSL , leaf-paradigms,text width=5em
                    [
                        ST Representation, leaf-paradigms, text width=9.5em
                        [\textbf{Jiang \textit{et al.}}~\cite{jiang2023self}{, } \textbf{Li \textit{et al.}}~\cite{li2023urban}{,} \textbf{Wan \textit{et al.}}~\cite{wan2021pre}{,} \textbf{Lin \textit{et al.}}~\cite{lin2022contrastive}{,} \textbf{Zhang \textit{et al.}}~\cite{zhang2023automated}{,} \\\textbf{Zhang \textit{et al.}}~\cite{zhang2023spatial}, modelnode-paradigms, text width=40em]
                    ]
                    [
                        ST Forecasting, leaf-paradigms, text width=9.5em
                         [\textbf{Xu \textit{et al.}}\cite{xu2020spatial}{, }\textbf{Liu \textit{et al.}}\cite{liu2022contrastive}{, }\textbf{Man \textit{et al.}}\cite{man2023w}{, }\textbf{Ranasinghe \textit{et al.}}~\cite{Ranasinghe_2022_CVPR}{,}\\
                         \textbf{Qu \textit{et al.}}\cite{qu2022forecasting}{,}
                         \textbf{Ji \textit{et al.}}\cite{ji2023spatio}{,}
                         \textbf{Wu \textit{et al.}}\cite{Wu_2023_CVPR}{,}
                         \textbf{Ranasinghe \textit{et al.}}~\cite{ranasinghe2023language}
                         , modelnode-paradigms, text width=40em]
                    ]
                    [
                        ST Recommendation, leaf-paradigms, text width=9.5em
                        [\textbf{Fu \textit{et al.}}~\cite{fu2024contrastive}{, }\textbf{Gao \textit{et al.}}~\cite{gao2023predicting}{, }\textbf{Zhou \textit{et al.}}~\cite{zhou2021contrastive}{, }\textbf{Gao \textit{et al.}}~\cite{gao2022self}{, }\textbf{Jiang \textit{et al.}}~\cite{jiang2023modeling}{, }\\\textbf{Zhou \textit{et al.}}~\cite{zhou2021self}, modelnode-paradigms,text width=40em]
                    ]
                    [
                        ST Other Tasks, leaf-paradigms, text width=9.5em
                        [\textbf{Sharma \textit{et al.}}~\cite{sharma2020self}{, }\textbf{Ayush \textit{et al.}}~\cite{ayush2021geography}{, }\textbf{Zhang \textit{et al.}}~\cite{zhang2020adaptive}{, }\textbf{Poppelbaum \textit{et al.}}~\cite{poppelbaum2022contrastive}{, }\\\textbf{Seong\textit{et al.}}~\cite{seong2024self}{, }\textbf{Jawed \textit{et al.}}~\cite{jawed2020self}{, }\textbf{Ma \textit{et al.}}~\cite{ma2020self}{, }\textbf{Si \textit{et al.}}~\cite{si2023dual}{, }\textbf{Tokmakov \textit{et al.}}~\cite{tokmakov2020unsupervised}{, }\\\textbf{Pei \textit{et al.}}~\cite{pei2023self}{, }\textbf{Sauder \textit{et al.}}~\cite{sauder2019self}, modelnode-paradigms,text width=40em]
                    ]
                ]
                [
                    Diffusion, leaf-others,text width=5em
                    [
                        ST Representation, leaf-others, text width=9.5em
                        [\textbf{Dai \textit{et al.}}~\cite{dai2024sadi}{, } \textbf{Tashiro \textit{et al.}}~\cite{tashiro2021csdi}, modelnode-others, text width=40em]
                    ]
                    [
                        ST Forecasting, leaf-others, text width=9.5em
                         [\textbf{Zhang \textit{et al.}}~\cite{zhang2023chattraffc}{, }\textbf{Hu \textit{et al.}}~\cite{hu2023towards}{, }\textbf{Liang \textit{et al.}}~\cite{liang2023dynamic}{, }\textbf{Zhou \textit{et al.}}~\cite{zhou2023towards}{, }\\\textbf{Wen \textit{et al.}}~\cite{wen2023diffstg}{, }\textbf{Zhu \textit{et al.}}~\cite{zhu2024difftraj}{, }\textbf{Ren \textit{et al.}}~\cite{ren2024tpllm}{, }\textbf{Lin \textit{et al.}}~\cite{lin2024specstg}{, }\textbf{Ling \textit{et al.}}~\cite{ling2024srndiff}{, }\\\textbf{Wei \textit{et al.}}~\cite{wei2024diff}{, }\textbf{Li \textit{et al.}}~\cite{li2023graph}{, }\textbf{Chu \textit{et al.}}~\cite{chu2024simulating}{, }\textbf{Hua \textit{et al.}}~\cite{hua2024weather}, modelnode-others, text width=40em]
                    ]
                    [
                        ST Recommendation, leaf-others, text width=9.5em
                        [\textbf{Qin \textit{et al.}}~\cite{qin2023diffusion}, modelnode-others,text width=40em]
                    ]
                    [
                        ST Other Tasks, leaf-others, text width=9.5em
                        [\textbf{Spetlik \textit{et al.}}~\cite{spetlik2024single}, modelnode-others,text width=40em]
                    ]
                ]
                [
                    Seq2Seq, leaf-other,text width=5em
                    [
                        ST Representation, leaf-other, text width=9.5em
                        [\textbf{Chang \textit{et al.}}~\cite{chang2018content}{, }\textbf{Li \textit{et al.}}~\cite{li2018deep}{, }\textbf{Zhang \textit{et al.}}~\cite{zhang2019deep}{, }\textbf{Liu \textit{et al.}}~\cite{liu2020representation}{, } \textbf{Taghizadeh \textit{et al.}}~\cite{taghizadeh2021meaningful}{, }\\ \textbf{Yao \textit{et al.}}~\cite{yao2022trajgat}{, } \textbf{Liu \textit{et al.}}~\cite{liu2022learning}{, } \textbf{Fu \textit{et al.}}~\cite{fu2020trembr}, modelnode-other, text width=40em]
                    ]
                    [
                        ST Forecasting, leaf-other, text width=9.5em
                         [\textbf{Wu \textit{et al.}}\cite{wu2020hierarchically}{, }\textbf{Jiang \textit{et al.}}~\cite{jiang2023pdformer}{, }\textbf{Jin \textit{et al.}}~\cite{jin2023trafformer}{, }\textbf{Chen \textit{et al.}}~\cite{chen2023prompt}{, }\textbf{Liu \textit{et al.}}~\cite{liu2017global}{, }\\\textbf{Wu \textit{et al.}}~\cite{wu2023transformerlight}{, }\textbf{Beladev \textit{et al.}}~\cite{beladev2023graphert}, modelnode-other, text width=40em]
                    ]
                    [
                        ST Recommendation, leaf-other, text width=9.5em
                        [\textbf{Yang \textit{et al.}}~\cite{yang2022getnext}{, }\textbf{Wu \textit{et al.}}~\cite{wu2023reason}{, }\textbf{Jiang \textit{et al.}}~\cite{jiang2023temporal}{, }\textbf{Halder \textit{et al.}}~\cite{halder2023capacity}{, }\textbf{Qin \textit{et al.}}~\cite{qin2022next}{, }\\\textbf{Halder \textit{et al.}}~\cite{halder2021transformer}, modelnode-other,text width=40em]
                    ]
                    [
                        ST Other Tasks, leaf-other, text width=9.5em
                        [\textbf{Fang \textit{et al.}}~\cite{fang20212}{, }\textbf{Wang \textit{et al.}}~\cite{wang2022deep}{, }\textbf{Kim \textit{et al.}}~\cite{kim2020anomalous}{, }\textbf{Wang \textit{et al.}}~\cite{wang2021learning}, modelnode-other,text width=40em]
                    ]
                ]
            ]
        \end{forest}
    }
    \caption{Taxonomy of existing studies based on \textit{generative techniques} that is composed of four kinds of techniques, including Large Language Models (LLMs), Self-Supervised Learning (SSL), Diffusion Models (Diffusion) and Sequence-to-Sequence (Seq2Seq), which include four spatial-temporal (ST) kinds of tasks with specific studies for each kind of task based on certain \textit{generative techniques}}
    \label{fig_tax}
\end{figure*}

%% file: eval.tex
\section{Applications}
\label{sec:eval}
In this section, we aim to provide illustration of applications. To provide clear illustrations on \textit{generation techniques} for several applications, namely \textbf{spatial-temporal representation learning}, \textbf{spatial-temporal forecasting}, \textbf{spatial-temporal recommendation} and \textbf{spatial-temporal clustering}, a comprehensive taxonomy shown in Figure~\ref{fig_tax} of existing studies has been developed based on generative techniques, encompassing four distinct categories of techniques: LLMs, SSL,Diffusion, and Seq2Seq models. Each of these categories provides unique approaches to address the challenges in spatial-temporal analysis. Within each category, specific studies have been conducted to tackle different types of spatial-temporal tasks by employing specific generative techniques tailored to those tasks. This taxonomy serves as a valuable framework for understanding and organizing the diverse body of research in the realm of spatial-temporal analysis, facilitating knowledge dissemination, and fostering further advancements in the field. The body of related works encompassing various application tasks within the field can be broadly categorized into different areas with corresponding datasets of different tasks, including spatial-temporal representation learning, spatial-temporal forecasting, and spatial-temporal recommendation, which are shown in Table ~\ref{tab:gen_method}. Each of these areas represents a distinct focus within spatial-temporal analysis, with numerous studies dedicated to advancing methodologies and techniques specifically tailored to address the challenges and demands of each task. By exploring the breadth and depth of research conducted in these areas, researchers gain valuable insights into the diverse approaches and innovative methodologies employed to tackle a wide range of spatial-temporal analysis tasks, ultimately advancing the field as a whole.

\subsection{Spatial-Temporal Representation Learning}
Representation learning is crucial for generating high-quality representations for various data types, including trajectories and spatial maps. These representations serve as foundations for downstream tasks, enhancing analysis efficiency. Although many studies have explored representation learning, the focus has mainly been on tailored representations for trajectories and spatial maps.

\subsubsection{Trajectories} \textbf{Non-Generative Methods:} Trajectories collected via GPS devices are used in map-matched form with road networks. RNN/LSTM-based methods excel in learning features from these sequences~\cite{hashemi2014critical,brakatsoulas2005map,quddus2007current,newson2009hidden}. Yang \textit{et al.}~\cite{yang2017neural} proposed a neural network model that learns representations of both aspects using RNN and GRU models. These models capture short-term and long-term relatedness, uncovering underlying patterns.

\noindent\textbf{Generative Methods:} With the advent of generative methods, recent studies~\cite{liu2022cstrm,jiang2023self,li2023self} have embraced self-supervised learning techniques to enhance supervised signals in trajectory analysis, thereby improving representation learning performance and enhancing trajectory similarities. For instance, Liu \textit{et al.}~\cite{liu2022cstrm} have introduced an innovative contrastive model that effectively learns trajectory representations by discerning differences at both the trajectory-level and point-level among trajectories. Furthermore, to mitigate the scarcity of training data, they propose a self-supervised approach to augment the available trajectory pairs for training. Jiang \textit{et al.}~\cite{jiang2023self} proposes to adopt self-supervised learning paradigm to model trajectory representation via capturing temporal regularities and travel semantics. Li \textit{et al.}~\cite{li2023self} adopts the contrastive learning paradigm to learn trajectory representations. For the research line of seq2seq~\cite{zhang2019deep,fu2020trembr,liu2020representation,li2018deep,yao2022trajgat,taghizadeh2021meaningful}, Zhang et al~\cite{zhang2019deep} employs a seq2seq-based method that takes vectors encompassing three types of semantic information as input. Fu \textit{et al.}~\cite{fu2020trembr} offers a novel approach with two key components. Firstly, it introduces Traj2Vec, an encoder-decoder model based on recurrent neural networks (RNNs). Traj2Vec encodes trajectory information by leveraging road networks to incorporate spatial and temporal properties. Secondly, Trembr introduces Road2Vec, a neural network model that learns embeddings for road segments within road networks. Road2Vec captures diverse relationships among road segments, preparing them for trajectory representation learning. Li \textit{et al.}~\cite{li2018deep} introduces seq2seq modeling to address trajectory similarity computation, handling challenges like non-uniform sampling rates and noisy sample points. CAPE~\cite{chang2018content} is a content-aware POI embedding model for personalized recommendations, training embedding vectors of POIs in a user's check-in sequence for proximity-based recommendations from historical check-in data. Tashiro \textit{et al}~\cite{tashiro2021csdi} proposes Conditional Score-based Diffusion models for Imputation (CSDI) to impute time series data via score-based diffusion models conditioned on observed data.

\subsubsection{Spatial Maps} \textbf{Non-Generative Methods:}  Several studies of non-generative methods have delved into the representation of spatial maps. HDGE~\cite{wang2017region} is built upon the flow graph via only human mobility data. ZE-Mob~\cite{yao2018representing} proposes to mine frequent patterns to further model relations of regions based on taxi and human trajectories. MV-PN~\cite{fu2019efficient} aims to capture the functions of regions based on POI data and human mobility data, which helps it have better performance on region representations. CGAL~\cite{zhang2019unifying} firstly constructs POI and mobility graph. Then CGAL model region representations via the unsupervised learning paradigm. This way makes CGAL to not only captures dependencies of intra-region but also captures dependencies of inter-region. MVURE~\cite{zhang2021multi} also proposes to capture region functions based on POI and human mobility data, whcih further captures correlations of regions. Lastly, MGFN~\cite{wu2022multi_graph} is a recently proposed method for region representation. It constructs a mobility graph and performs message passing to generate region embedding.

\noindent\textbf{Generative Methods:} Recent studies~\cite{zhang2023automated,zhang2023spatial,wan2021pre,lin2022contrastive} via generative methods have explored representation learning of spatial maps using self-supervised learning. These frameworks aim to provide general embeddings for various downstream tasks such as trajectory prediction, traffic prediction, crime prediction, and house price prediction. One approach, proposed by Wan \textit{et al.}~\cite{wan2021pre}, is Time-Aware Location Embedding (TALE) pretraining. TALE utilizes the CBOW framework to enhance embedding vectors of locations by incorporating temporal information. It introduces a unique temporal tree structure for extracting temporal details during Hierarchical Softmax computation. TALE's efficacy is validated through location classification, prediction of location visitor flow, and user's next location prediction tasks. Lin \textit{et al.}~\cite{lin2022contrastive} proposes a framework, namely MMTEC via maximizing encoding of multi-view trajectory entropy, thereby providing extensive trajectory embeddings. They address biases in pre-trained trajectory embeddings through a preliminary task, making the embeddings valuable for diverse downstream applications. Zhang \textit{et al.}~\cite{zhang2023automated} presents AutoST (Automated Spatio-Temporal), which employs contrastive learning on a heterogeneous region graph created from multiple data sources. AutoST's graph neural architecture effectively captures dependencies between regions from different views based on POI data, mobility data to capture region functions, mobility flow patterns, and positional relations regions. Additionally, Zhang \textit{et al.}~\cite{zhang2023spatial} proposes GraphST, a model focused on spatial-temporal graph learning for effective self-supervised learning. GraphST utilizes adversarial contrastive learning to distill essential self-supervised information from multiple views, improving spatial-temporal graph augmentation. Another recent work~\cite{li2023urban} leverages OpenStreetMap (OSM) buildings' footprints in conjunction with points of interest (POIs) to acquire region representations via self-supervised learning paradigm. This approach is motivated by the fact that buildings' shapes, spatial distributions, and properties are intricately connected to various urban functions.
Another recent work~\cite{yan2023urban} harnesses the capabilities of Large Language Models (LLMs) and introduce UrbanCLIP, a pioneering framework that seamlessly incorporates the knowledge of textual modality into urban imagery profiling. Zhang \textit{et al.}~\cite{zhang2023spatio} proposes to augment region representation learning paradigm via LLMs. To be specific, Liu \textit{et al.}~\cite{liu2022learning} presents a pioneering approach called Trajectory-aware Transformer for Video Super-Resolution (TTVSR), which reframed video frames into pre-aligned trajectories comprising consecutive visual tokens. When processing a query token, self-attention is exclusively focused on pertinent visual tokens along the spatial-temporal trajectories. For diffusion models on representation learning, Dai \textit{et al.}~\cite{dai2024sadi} introduces a framework called Similarity-Aware Diffusion Model-Based Imputation (SADI), which harnesses the power of the diffusion model. By leveraging the inherent properties of the diffusion model, SADI offers a robust and innovative approach to imputing incomplete temporal electronic health record data.

\begin{table*}[!t]
\scriptsize
\caption{Related works in different application tasks, namely spatial-temporal representation learning, spatial-temporal forecasting, and spatial-temporal recommendation. We provide detailed information on the names of journals/conferences, the specific generative techniques used, the particular application sub-tasks addressed, and the datasets used for evaluation.}
\setlength{\tabcolsep}{2.7mm}{
\resizebox{1.0\linewidth}{!}{
\begin{tabular}{l|c|c|c|c}
\toprule
\rowcolor{lightgray} \textbf{Method} & \textbf{Source} & \textbf{Technique} & \textbf{Sub-task} & \textbf{Datasets} \\ \hline
\rowcolor{lightgray}\multicolumn{5}{l}{\textbf{Spatial-temporal Representation Learning}} \\ \hline
\rowcolor{lightred} UrbanCLIP~\cite{yan2023urban} & \textit{WWW 2024}  & LLM & Region Profiling & Beijing, Shanghai, Guangzhou, Shenzhen \\ \hline
\rowcolor{lightred} STLLM~\cite{zhang2023spatio} & \textit{ICLR 2024}  & LLM & Spatio-temporal Representation & CHITaxi, NYCBike, NYCTaxi, CHICrime, NYCCrime, CHIHouse, NYCHouse \\ \hline
\rowcolor{lightpurple} Liu \textit{et al.}~\cite{liu2022cstrm} & \textit{COMPUT COMMUN}  & SSL & Trajectory Similarity & Porto, TaxiBJ \\ \hline
\rowcolor{lightpurple} START~\cite{jiang2023self} & \textit{ICDE 2023}  & SSL & Trajectory Representation & Porto, Beijing \\ \hline
\rowcolor{lightpurple} TrajRCL~\cite{li2023self} & \textit{FUTURE GENER COMP SY}  & SSL & Trajectory Representation & Porto, T-Drive \\ \hline
\rowcolor{lightpurple} AutoST~\cite{zhang2023automated}& \textit{WWW 2023}  & SSL & Spatio-temporal Representation & Census, Taxi, POI, CHICrime, NYCCrime, CHIHouse, NYCHouse \\ \hline
\rowcolor{lightpurple} GraphST~\cite{zhang2023spatial}& \textit{ICML 2023}  & SSL & Spatio-temporal Representation & CHITaxi, NYCBike, NYCTaxi, CHICrime, NYCCrime, CHIHouse, NYCHouse \\ \hline
\rowcolor{lightgreen} DiffPose~\cite{feng2023diffpose} & \textit{ICCV 2023}  & Diffusion & Pose Representation & PoseTrack2017, PoseTrack2018, PoseTrack21 \\ \hline
\rowcolor{lightyellow} TrajGAT ~\cite{yao2022trajgat}& \textit{KDD 2022}  & Seq2Seq & Trajectory Similarity & DiDi-Xian \\ \hline
\rowcolor{lightyellow} TTVSR~\cite{liu2022learning}& \textit{CVPR 2022}  & Seq2Seq & Video Super-resolution & REDS, Vimeo-90K \\ \hline
\hline
\rowcolor{lightgray} \multicolumn{5}{l}{\textbf{Spatial-temporal Forecasting}} \\ \hline
\rowcolor{lightred} PromptCast~\cite{xue2023promptcast}  & \textit{TKDE}  & LLM & Spatial-temporal Forecasting & City Temperature, Electricity Consumption, SafeGraph  \\ \hline
\rowcolor{lightred} Yu \textit{et al.}~\cite{yu2023temporal}  & \textit{Arxiv 2023}  & LLM & Financial Forecasting & NASDAQ-100  \\ \hline
\rowcolor{lightred} FPT~\cite{zhou2024one} & \textit{NeurIPS 2023}  & LLM & Long-term Forecasting & ETTh1/h2/m1/m2, Weather, Electricity, Traffic, ILI  \\ \hline
\rowcolor{lightred} LLMTIME~\cite{gruver2024large} & \textit{NeurIPS 2023}  & LLM & Spatial-temporal Forecasting & Istanbul Traffic, TSMC Stock, Turkey Power  \\ \hline
\rowcolor{lightred} ST-LLM~\cite{liu2024spatial} & \textit{Arxiv 2024}  & LLM & Traffic Prediction & NYCTaxi~\cite{liu2020dynamic}, CHIBike~\cite{wang2021libcity}\\ \hline
\rowcolor{lightred} UniST~\cite{yuan2024unist}  & \textit{Arxiv 2024}  & LLM & Traffic Prediction & TaxiBJ, Cellular, NYCTaxi~\cite{liu2020dynamic}, T-Drive~\cite{pan2019urban}, Crowd, TrafficXX\\ \hline
\rowcolor{lightred} GPT4MTS~\cite{jia2024gpt4mts} & \textit{AAAI 2024}  & LLM & Spatial-temporal Forecasting &  Global Database of Events, Language, and Tone  \\ \hline
\rowcolor{lightred} Time-LLM~\cite{jin2023time} & \textit{ICLR 2024}  & LLM & Long-term Forecasting & ETTh1/h2/m1/m2, Weather, Electricity, Traffic, ILI  \\ \hline
\rowcolor{lightred} TEMPO~\cite{cao2023tempo} & \textit{ICLR 2024}  & LLM & Long-term Forecasting & ETTh1/h2/m1/m2, Weather, Electricity, TETS, GDELT  \\ \hline
\rowcolor{lightpurple} STTN~\cite{xu2020spatial} & \textit{Arxiv 2020}  & SSL & Traffic
Prediction  & PEMS-BAY, PeMSD7(M)  \\ \hline
\rowcolor{lightpurple} Liu \textit{et al.}~\cite{liu2022contrastive} & \textit{SIGSPATIAL 2022}  & SSL & Traffic
Prediction  &  PeMSD4,  PeMSD8  \\ \hline
\rowcolor{lightpurple} SVT~\cite{Ranasinghe_2022_CVPR} & \textit{CVPR 2022}  & SSL & Action
Recognition  & Kinetics-400, UCF-101, HMDB51, SSv2  \\ \hline
\rowcolor{lightpurple} UrbanSTC~\cite{qu2022forecasting}& \textit{TKDE}  & SSL & Traffic Prediction & TaxiBJ, BikeNYC  \\ \hline
\rowcolor{lightpurple} W-MAE~\cite{man2023w} & \textit{Arxiv 2023}  & SSL & Weather Forecasting & ERA5  \\ \hline
\rowcolor{lightpurple} ST-SSL~\cite{ji2023spatio} & \textit{AAAI 2023}  & SSL & Traffic Prediction & NYCBike, NYCtaxi, BJTaxi  \\ \hline
\rowcolor{lightpurple} Wu \textit{et al.}~\cite{Wu_2023_CVPR} & \textit{AAAI 2023}  & SSL & Point Cloud Segmentation & KITTI, NuScene, SemanticKITTI, SemanticPOSS   \\ \hline
\rowcolor{lightpurple} LSS-A~\cite{ranasinghe2023language} & \textit{NeurIPS 2023}  & SSL & Action Recognition & Kinetics400, UCF-101, HMBD-51   \\ \hline
\rowcolor{lightgreen} ChatTraffic~\cite{zhang2023chattraffc} & \textit{Arxiv 2023}  & Diffusion & Traffic Prediction & Beijing~\cite{zhang2023chattraffc} \\ \hline
\rowcolor{lightgreen} USTD~\cite{ hu2023towards} & \textit{Arxiv 2023}  & Diffusion & Spatial-temporal Forecasting &  PEMS-03~\cite{song2020spatial}, PEMS-BAY~\cite{li2017diffusion}, AIR-BJ~\cite{yi2018deep}, AIR-GZ~\cite{yi2018deep} \\ \hline
\rowcolor{lightgreen} DVGNN~\cite{liang2023dynamic} & \textit{Arxiv 2023}  & Diffusion & Spatial-temporal Forecasting & PeMS08~\cite{song2020spatial}, Los-loop, T-Drive~\cite{pan2019urban}, FMRI-3,4,13 \\ \hline
\rowcolor{lightgreen} KSTDif~\cite{zhou2023towards} & \textit{SIGSPATIAL 2023}  & Diffusion & Traffic Prediction & New York City, Beijing, Washington, Baltimore\\ \hline
\rowcolor{lightgreen} DiffSTG~\cite{wen2023diffstg} & \textit{SIGSPATIAL 2023}  & Diffusion & Spatial-temporal Forecasting & PeMS08~\cite{song2020spatial}, AIR-BJ~\cite{yi2018deep}, AIR-GZ~\cite{yi2018deep} \\ \hline
\rowcolor{lightgreen} Diff-Traj~\cite{zhu2024difftraj} & \textit{NeurIPS 2023}  & Diffusion & Trajectory Generation & Chengdu, Xi’an \\ \hline
\rowcolor{lightgreen} TPLLM~\cite{ren2024tpllm} & \textit{INT J GEOGR INF SCI}  & Diffusion & Mobility Modeling & T-Drive~\cite{pan2019urban}, Geo-life~\cite{hasan2015location}\\ \hline
\rowcolor{lightgreen} SpecSTG~\cite{lin2024specstg} & \textit{Arxiv 2024}  & Diffusion & Traffic Prediction & PeMS04,08~\cite{song2020spatial}\\ \hline
\rowcolor{lightgreen} SRNDiff~\cite{ling2024srndiff} & \textit{Arxiv 2024}  & Diffusion & Rainfall Prediction & Nimrod-uk-1km \\ \hline
\rowcolor{lightgreen} Diff-RNTraj~\cite{wei2024diff} & \textit{Arxiv 2024}  & Diffusion & Trajectory Generation & Porto, Chengdu \\ \hline
\rowcolor{lightgreen} GCRDD~\cite{li2023graph} & \textit{ADMA 2023}  & Diffusion & Traffic Prediction & MeTR-LA, PEMS-BAY \\ \hline
\rowcolor{lightgreen} TrajGDM~\cite{chu2024simulating} & \textit{GIS 2024}  & Diffusion & Trajectory Generation & Porto, Chengdu \\ \hline
\rowcolor{lightgreen} Hua \textit{et al.}~\cite{hua2024weather} & \textit{Arxiv 2024}  & Diffusion & Weather Forecasting & WeatherBench~\cite{rasp2020weatherbench} \\ \hline
\rowcolor{lightyellow} PDFormer~\cite{jiang2023pdformer} & \textit{AAAI 2023}  & Seq2Seq & Traffic Prediction & PeMS04,07,08~\cite{song2020spatial}, NYCTaxi~\cite{liu2020dynamic}, CHIBike~\cite{wang2021libcity}, T-Drive~\cite{pan2019urban}  \\ \hline
\rowcolor{lightyellow} Trafformer~\cite{jin2023trafformer} & \textit{AAAI 2023}  & Seq2Seq & Traffic Prediction & METR-LA, PEMS-Bay.  \\ \hline
\rowcolor{lightyellow} Chen \textit{et al.}~\cite{chen2023prompt} & \textit{IJCAI 2023}  & Seq2Seq & Weather Forecasting & AvePRE~\cite{liu2017global}, SurTEMP~\cite{liu2017global}, SurUPS~\cite{liu2017global} \\ \hline
\rowcolor{lightyellow} TransformerLight~\cite{wu2023transformerlight} & \textit{KDD 2023}  & Seq2Seq & Sequence Modeling & Jinan~\cite{zheng2019learning}, Hangzhou~\cite{zheng2019learning}, New York~\cite{zheng2019learning} \\ \hline
\rowcolor{lightyellow} GraphERT~\cite{beladev2023graphert} & \textit{CIKM 2023}  & Seq2Seq & Temporal Classification & Facebook
wall posts, Enron, Reddit  \\ \hline\hline
\rowcolor{lightgray} \multicolumn{5}{l}{\textbf{Spatial-temporal Recommendation}} \\ \hline
\rowcolor{lightred}LLMmove~\cite{feng2024move} & \textit{Arxiv 2024}  & LLM & POI Recommendation &NYC~\cite{yang2014modeling}, TKY~\cite{yang2014modeling}\\ \hline
\rowcolor{lightred}Li \textit{et al.}~\cite{li2024large} & \textit{SIGIR 2024}  & LLM & POI Recommendation &NYC~\cite{yang2014modeling}, TKY~\cite{yang2014modeling}, Gowalla-CA \\ \hline
\rowcolor{lightpurple}SLS-REC~\cite{fu2024contrastive} & \textit{ES 2024}  & SSL & POI Recommendation &Foursquare~\cite{yang2014modeling}, Gowalla \\ \hline
\rowcolor{lightpurple}Gao \textit{et al.}~\cite{gao2023predicting} & \textit{TKDE 2023}  & SSL & Human Trajectory Prediction &NYC, TKY, Los Angeles, Houston \\ \hline
\rowcolor{lightpurple}CTLTR~\cite{zhou2021contrastive} & \textit{TIST 2021}  & SSL & Tour Recommendation &Edinburgh, Glasgow, Osaka, Toronto \\ \hline
\rowcolor{lightpurple}LSPSL~\cite{jiang2023modeling} & \textit{TKDD 2023}  & SSL & POI Recommendation &NYC, TKY \\ \hline
\rowcolor{lightpurple}SML~\cite{zhou2021self} & \textit{KBS 2021}  & SSL & Next Location Prediction &NYC, Singapore, Houston, California \\ \hline
\rowcolor{lightgreen}Diff-POI~\cite{qin2023diffusion} & \textit{TOIS 2023}  & Diffusion & POI Recommendation &Gowalla,  Singapore, TKY, NYC \\ \hline
\rowcolor{lightyellow}TLR-M~\cite{halder2021transformer} & \textit{KDDM 2021}  & Seq2Seq &POI Recommendation  &NYC, TKY \\ \hline
\rowcolor{lightyellow}GETNext~\cite{yang2022getnext} & \textit{SIGIR 2022}  & Seq2Seq &POI Recommendation  &NYC, TKY, CA \\ \hline
\rowcolor{lightyellow}HAT~\cite{wu2023reason} & \textit{IT MM 2023}  & Seq2Seq &POI Recommendation  &BJ, SH, NJ, CD \\ \hline
\rowcolor{lightyellow}TGAT~\cite{jiang2023temporal} & \textit{IF Systems 2023}  & Seq2Seq &POI Recommendation  &NYC, TKY \\ \hline
\rowcolor{lightyellow}CAFPR~\cite{halder2023capacity} & \textit{Applied SC 2023}  & Seq2Seq &POI Recommendation  &TKY, CA, Budapest~\cite{chen2020personalized}, Melbourne~\cite{chen2020personalized}, Magic k~\cite{chen2020personalized} \\ \hline
\rowcolor{lightyellow}AutoMTN~\cite{qin2022next} & \textit{SIGIR 2022}  & Seq2Seq &POI Recommendation  &NYC, TKY \\ \hline

\bottomrule
\end{tabular}}}
\label{tab:gen_method}
\end{table*}

\subsection{Spatial-Temporal Forecasting} Spatial-temporal forecasting methods target learning a function that can map input features to predicted results via a well-explored training set. In spatial-temporal applications, the input and output variables can encompass different types of spatial-temporal (ST) data instances, leading to diverse problem formulations for predictive learning. Spatial-temporal applications include traffic prediction, crime prediction, climate prediction and so on. In the following, we will delve into several commonly encountered predictive learning problems in ST applications based on the type of ST data instance employed as input variables.

\input{pre}

\subsection{Spatial-Temporal Recommendation}
Spatial-temporal recommendation refers to the task of recommending relevant POI or locations to users according to their spatial-temporal context. It leverages the availability of location and time information to deliver personalized recommendations that consider both the user's preferences and the specific spatial-temporal context. By analyzing patterns and correlations in spatial and temporal data, such as user movements, check-in histories, and time-dependent preferences, spatial-temporal recommendation systems can provide valuable suggestions for various applications, including travel planning, event recommendations, and location-based advertising. These systems aim to enhance the user experience by offering timely and location-specific recommendations that align with the user's current context and preferences.

\subsubsection{Point} \textbf{Non-Generative Methods:} Earlier studies~\cite{li2020group,huang2019attention,chang2018content,wu2020personalized}  For example, Huang \textit{et al.}~\cite{huang2019attention} introduces a framework called the Spatial-Temporal Long and Short-Term Memory (ST-LSTM) which enhances the modeling of spatial and temporal information by incorporating spatial-temporal contextual data into the LSTM network at every step. Another work~\cite{chang2018content} proposes the pioneering content-aware POI embedding model that leverages textual content to capture valuable information regarding the distinctive features of a Point of Interest (POI). Wu \textit{et al.}~\cite{wu2020personalized} merges the long-term and short-term preferences by utilizing a user-based linear combination unit. This unit enables us to learn personalized weights for different users, allowing them to assign varying importance to different components.

\noindent\textbf{Generative Methods:} Recent studies like ~\cite{feng2024move,li2024large} for POI recommendation conducted via LLMs are described in detail as follows: Feng \textit{et al.}~\cite{feng2024move} develops innovative prompting strategies and carry out empirical studies to evaluate the predictive ability of LLMs, such as ChatGPT, in anticipating a user's next check-in via capturing various crucial aspects of human movement patterns, such as user preferences regarding location, spatial proximity, and sequential transitions. Li \textit{et al.}~\cite{li2024large} proposes to preserve heterogeneous location-based social networking data in its original format, thereby preventing the loss of valuable contextual information. Another research line via diffusion models for POi recommendation is presented as follows: Qin \textit{et al.}~\cite{qin2023diffusion} introduces a novel diffusion-based model that leverages user spatial preferences for generating next Point of Interest (POI) recommendations via sampling techniques to capture the user's spatial preferences and incorporate them into the recommendation process. For the research line of the self-supervised learning paradigm~\cite{fu2024contrastive,zhou2021self,zhou2021contrastive,gao2022self,jiang2023modeling}, to be specific, Jiang \textit{et al.}~\cite{jiang2023modeling} adopts the self-supervised learning paradigm to model long-term preferences and short-term preferences to enhance the accuracy and effectiveness of POI recommendation systems. Wang \textit{et al.}~\cite{wang2023exploring} captures long-term preferences and short-term preferences on visiting positions of users via a self-supervised learning mechanism by multi-feature embedding via integrating trajectory graph representation. Gao \textit{et al.}~\cite{gao2022self} introduces a contrastive learning mechanism via two steps and augment four trips designed to capture the uncertainties associated with planning trips, thereby enhancing the overall effectiveness of the recommendation system. For the research line of seq2seq~\cite{yang2022getnext,halder2021transformer,qin2022next,halder2023capacity,jiang2023temporal,wu2023reason}, to be specific, Yang \textit{et al.}~\cite{yang2022getnext} introduces a global trajectory flow map via the user-guided way and a novel model called Graph Enhanced Transformer (GETNext) to leverage compressive external signals and mitigate the challenges associated with the cold start problem. Qin \textit{et al.}~\cite{qin2022next} introduces AutoMTN (Auto-correlation Enhanced Multi-modal Transformer Network), a novel model designed specifically for the next Point of Interest (POI) recommendation. AutoMTN leverages auto-correlation techniques to enhance its multi-modal transformer network.

\subsubsection{Time Series} \textbf{Non-Generative Methods:} Former studies~\cite{choe2021recommendation,bedi2021session,zhou2020cnn,wang2022ngcu}. For example, Zhou \textit{et al.}~\cite{zhou2020cnn} developes a novel neural network structure called DP-CRNN algorithm, which aims to extract and emphasize the fusion of semantic and sequential features from patient inquiries. Wang \textit{et al.}~\cite{wang2022ngcu} adopts RNN, LSTM, and GRU to perform comparative experiments based on three distinct datasets comprising air quality, Hang Seng Index, and gold future prices to substantiate the generalization of NGCU.

\noindent\textbf{Generative Methods:} For the research line of self-supervised learning~\cite{gao2023predicting}, ~\cite{gao2023predicting} proposes to address the issue of label sparsity by enhancing the learning of latent relationships among item features via introducing a novel data augmentation technique with SSL.

\subsection{Spatial-Temporal Clustering}
Clustering groups similar data points together to discover inherent structures within a dataset. Researchers have explored this research line for a long time. Detailed information is shown as follows:

\subsubsection{Points}

\textbf{Non-Generative Methods:}  Clustering aims to identify hot-spots, areas with elevated density of ST points~\cite{atluri2018spatio}. It detects outbreaks or social movements with concentrated events in space and time~\cite{gomide2011dengue,kulldorff2005space}. Dynamic Auto-encoder~\cite{mrabah2019deep} balances clustering and reconstruction, preserving spatial topology.

\noindent\textbf{Generative Methods:} Recent studies~\cite{sharma2020self, sauder2019self} have focused on adopting the self-supervised learning paradigm for point clouds. In particular, Sharma \textit{et al.}~\cite{sharma2020self} proposes to adopt a cover-tree, during each layer of it, subsets of the point clouds are grouped within balls of varying radii, via self-supervised learning paradigm to divide point clouds hierarchical. On the other hand, Sauder \textit{et al.}~\cite{sauder2019self} presents a self-supervised learning task specifically designed for deep learning with raw point cloud data. This task involves training a neural network to reconstruct point clouds while randomly rearranging parts of the point clouds. Simultaneously, the neural network learns representations that effectively capture the semantic properties of the point clouds. A recent work~\cite{pei2023self} uses temporal variation in vehicle emissions for high-purity clustering. They employ a pretrained BiLSTM network for initial temporal representation and utilize a self-supervised mechanism with GCN for feature clustering. Joint optimization refines representation and clustering progressively~\cite{pei2023self}.

\subsubsection{Trajectories} \textbf{Non-Generative Methods:} The method proposed in~\cite{lee2007trajectory} divides individual trajectories into sets of line segments and then groups together similar line segments to form clusters. The key benefit of this framework is its capability to uncover shared sub-trajectories within a trajectory database. Leveraging this partition-and-group framework, this method has developed an effective trajectory clustering algorithm known as TRACLUS. Kumaran \textit{et al.}~\cite{kumaran2018video} introduces a high-level representation of object trajectories by utilizing the color gradient form.

\noindent\textbf{Generative Methods:} In a recent work~\cite{tokmakov2020unsupervised}, the adaptation of two effective objectives, instance recognition and local aggregation, to the video domain is proposed. The local aggregation approach involves iteratively clustering videos in a network's feature space and updating the network using a non-parametric classification loss to align with the clusters. Although promising performance is observed, qualitative analysis reveals a limitation in capturing motion patterns, as the videos are primarily grouped based on appearance. To address this, the method incorporates heuristic-based IDT descriptors, which aims to encode motion patterns of videos and regarded as an unsupervised prior in the local aggregation algorithm for each iteration, enabling the formation of clusters in the IDT space. Another recent work~\cite{si2023dual} proposes a dual Self-supervised Deep Trajectory Clustering (SDTC) method that jointly optimizes trajectory representation and clustering. Firstly, this method leverages the BERT model to learn spatial-temporal mobility patterns and incorporate them into the embeddings of location IDs. Secondly, this method fine-tunes the BERT model to enhance the cluster-friendliness of trajectory representations by introducing a dual self-supervised cluster layer. This layer improves intra-cluster similarities and promotes inter-cluster dissimilarities, leading to more effective trajectory clustering. Fang \textit{et al.}~\cite{fang20212} presents E2DTC, an end-to-end deep trajectory clustering framework that employs self-training. E2DTC eliminates the need for manual feature extraction operations and can be seamlessly applied to perform trajectory clustering analytics on any trajectory dataset.

\subsubsection{Time Series} \textbf{Non-Generative Methods:}  Time series clustering has been explored extensively. Richard~\cite{richard2020autoencoder} combines a convolutional autoencoder and k-medoids for clustering, enhancing feature extraction and dimensionality reduction. Thinsungnoen~\cite{thinsungnoen2018deep} focuses on high-dimensional time series like ECGs, using Deep Autoencoder Networks (DANs) for representative feature extraction. Genetic algorithms optimize the DAN structure for clustering. Tavakoli \textit{et al.}~\cite{tavakoli2020autoencoder} and Li \textit{et al.}~\cite{li2022autoshape} propose autoencoder-based models to capture nonlinear features in time series data.

\noindent\textbf{Generative Methods:} Recent studies~\cite{ma2020self,jawed2020self,seong2024self,poppelbaum2022contrastive} have embraced the self-supervised learning paradigm for time series clustering. In self-supervised learning, the goal is to leverage data augmentation techniques to enable deep neural networks to extract relevant information. For instance, Poppelbaum \textit{et al.}~\cite{poppelbaum2022contrastive} proposes a novel self-supervised learning paradigm to analyze time-series data, specifically designed via SimCLR contrastive learning paradigm. They introduce innovative data augmentation techniques, with a particular focus on time-series data, and investigate their impact on the prediction task.

\subsubsection{Spatial Maps} \textbf{Non-Generative Methods:} Former studies~\cite{li2020automatic} To be specific, Li \textit{et al.}~\cite{li2020automatic,pan2018spatial} introduced a two-branch CNN that utilizes automatic clustering for efficient processing. In the first branch, the CNN subdivides the Hyperspectral Image (HSI) pixels into smaller classes through clustering, thereby reducing intra-class spectral variation. Pan \textit{et al.}~\cite{pan2018spatial} introduces a novel approach called Spatial CNN (SCNN) that extends the conventional deep convolutions via layer-by-layer paradigm to convolutions in feature maps via the slice-by-slice paradigm, which enables inter-pixel communication across rows and columns within a layer, facilitating effective message passing.

\noindent\textbf{Generative Methods:} In a recent work~\cite{wang2022deep}, the method begins by employing a sequence-to-sequence based network for pretraining. This enables the method to obtain initial trajectory vectors. Subsequently, we optimize the learned representations and cluster centroids simultaneously within a unified framework based om spatial maps. Another recent work~\cite{zhang2020adaptive} introduces a novel approach that incorporates a graph convolutional network design to capture information cues for characterizing both intra- and inter-image correspondence. Furthermore, this paper develops an attention-based graph clustering algorithm that discriminates common objects from salient foreground objects in an unsupervised manner. To achieve seamless integration, this paper proposes a unified framework with an encoder-decoder structure, enabling joint training and optimization of the graph convolutional network, attention graph cluster, and co-saliency detection decoder in an end-to-end fashion. Another papers based on up-to-date generative methods~\cite{ayush2021geography,zhang2019self} demonstrate that contrastive and supervised learning exhibit a notable discrepancy on conventional benchmarks due to their distinct characteristics. To bridge this gap, Ayush \textit{et al.}~\cite{ayush2021geography} introduces innovative training methods that capitalize on the spatial-temporal structure of remote sensing data. By utilizing spatially aligned images collected over time, the positive temporal pairs for contrastive learning are built. Additionally, Ayush \textit{et al.}~\cite{ayush2021geography} leverages geo-location information to design pre-text tasks that enhance the learning process. Another research line via LLMs is illustrated in the following. Guo \textit{et al.}~\cite{guo2023point} introduces Point-LLM, the pioneering 3D LLM that incorporates 3D multi-modal instructions. Through efficient fine-tuning methods, Point-LLM seamlessly integrates the semantics of Point-Bind into pre-trained LLMs. Hong \textit{et al.}~\cite{hong20233d} introduce a novel approach of incorporating the 3D world into large language models. They achieve this by creating a new category of models called 3D-LLMs and designing three distinct prompts.

\subsection{Other Tasks} 
\textbf{Non-Generative Methods: } Earlier studies on other spatial-temporal tasks like anomaly detection~\cite{zhang2023online,chen2013iboat,song2018anomalous,nanduri2016anomaly} and relationship mining~\cite{qi2021chinese} adopted CNN- or RNN-based methods to model anomalous trajectories or spatial-temporal relationships. To be specific, Zhang \textit{et al.}~\cite{zhang2023online} adopts LSTM to model features of sub-trajectories; based on these features, reinforcement learning is adopted to detect anomalous trajectories. Chen \textit{et al.}~\cite{chen2013iboat} investigates online detection of anomalous trajectories using the isolation-based method. However, this method relies on manually-set parameters and is evaluated on a small dataset, limiting its generalization. Another recent work~\cite{zhang2020continuous} proposes trajectory similarity calculation using Frechet distance. 

\noindent\textbf{Generative Methods: } Recent studies~\cite{wang2021learning,kim2020anomalous,spetlik2024single} on pattern mining and anomalous trajectory detection via generative methods are presented in details. Spetlik \textit{et al.}~\cite{spetlik2024single} proposes to adopt a diffusion model-based method to conduct trajectory recovery. Wang \textit{et al.}~\cite{wang2021learning} introduces a fusion of a recurrent neural network-based seq2seq model to conduct time series pattern mining, an attention mechanism, and an enhanced set of features extracted using dynamic time warping and zigzag peak valley indicators. Kim \textit{et al.}~\cite{kim2020anomalous} presents a Seq2Seq Auto-Encoder-based model that adeptly captures unique high-quality features and movement characteristics from the trajectories.

%% file: pre.tex
\subsubsection{Points}

\textbf{Non-Generative Methods:} Earlier studies based on non-generative methods are shown as following: DASTGNN~\cite{guo2021learning} incorporates self-attention and dynamic graph convolutions. DMSTGCN~\cite{han2021dynamic} enhances graph convolutions with dynamic geographical and temporal data. DSTAGNN~\cite{lan2022dstagnn} uses a dynamic spatial-temporal aware graph and improved multi-head attention. FOGS~\cite{rao2022fogs} builds an association graph based on spatial-temporal dynamics. STGNCDE~\cite{choi2022graph} uses graph neural controlled differential equations for traffic prediction. STSHN~\cite{xia2022spatial} uses hypergraph connections for spatial message transfer between regions.

\noindent\textbf{Generative Methods:} More recently, with the success of generative methods in many domains,
STGCL~\cite{liu2022contrastive} presents the first systematic exploration of integrating contrastive learning into traffic data analysis. Specifically, the study outlines two potential schemes for this integration and proposes two feasible and efficient designs for contrastive tasks, which are executed at the node or graph level.
STSSL~\cite{Wu_2023_CVPR} utilizes positive pairs in both spatial and temporal domains, incorporating a point-to-cluster learning strategy for spatial object differentiation and a cluster-to-cluster approach for exploiting temporal correspondences through unsupervised object tracking.

Transformer and diffusion architectures are progressively being recognized as powerful methods.
STTN~\cite{xu2020spatial} proposes Spatial-Temporal Transformer Networks that leverage dynamical directed spatial dependencies and long-range temporal dependencies for accurate long-term traffic flow forecasting. 
By stacking two layers of Transformer to represent spatial-temporal links across spaces and time, STrans~\cite{wu2020hierarchically} investigates the sparse crimes. For the aggregation of spatial and temporal information, self-attention with query/key transformations is used.
TFormer~\cite{jin2023trafformer} combines temporal and spatial data into a single transformer-style model, allowing nodes at different timestamps to interact in a single step to capture intricate spatial-temporal relationships.
PDFormer~\cite{jiang2023pdformer} employs a spatial self-attention module to capture changing spatial inter-dependencies. It also utilizes graph masking matrices to emphasize spatial dependencies from short- and long-range perspectives for precise traffic flow prediction.
Furthermore, TPLLM~\cite{ren2024tpllm} leverages pretrained LLMs to overcome the challenges of data scarcity in intelligent transportation systems. Utilizing a combination of CNN and GCN for feature extraction and a Low-rank adaptation fine-tuning method, TPLLM demonstrates robust performance in both full-sample and few-shot scenarios on real-world traffic datasets.
There is also an increasing number of probabilistic forecasting predictive models based on diffusion.
DiffSTG~\cite{wen2023diffstg} integrates the spatial-temporal learning capabilities of ST-GNNs with the uncertainty quantification features of diffusion models, leading to more accurate and reliable predictions. 
DVGNN~\cite{liang2023dynamic} integrates a generative model in the decoder stage to discover causal relationships, which allows the model to better handle the uncertainty and noise inherent in real-world traffic prediction.
UTSD~\cite{hu2023towards} utilizes diffusion models to model complex distributions and manage uncertainties inherent in traffic data. 
SpecSTG~\cite{lin2024specstg} extends the idea of probabilistic forecasting to fully capture the uncertainties and risks associated with future traffic patterns, in which the key advantage is its transformation of the learning process into the spectral domain.

\subsubsection{Trajectories} 

\textbf{Non-Generative Methods:}  Former studies~\cite{ASTGCN,BaiTCN2018,shleifer2019incrementally,han2022lst,huang2019dsanet,DCRNN,STGCN} focus on capturing trajectory patterns via Temporal Convolutional Networks combing with attention mechanisms for trajectories prediction. For example, STResNet~\cite{zhang2017deep} utilizes a residual neural network framework to capture the temporal characteristics of crowd traffic and dynamically aggregates the output of three residual neural networks.
DCRNN~\cite{DCRNN}, employs a diffusion convolutional recurrent neural network (RNN) with a fusion process to model spatial-temporal dependencies.

\noindent\textbf{Generative Methods:} In recent study,
SPGCL~\cite{li2022mining} advances spatio-temporal learning by dynamically optimizing graph structures through a self-paced strategy. It distinguishes informative relations by enhancing the margin between positive and negative neighbors. In each iteration, neighborhoods selectively integrate new nodes with the highest affinity scores, thereby refining and expanding the graph. 
Following a different approach, 
UrbanSTC~\cite{qu2022forecasting} develops a contrastive self-supervision method designed to predict fine-grained urban flows by integrating diverse spatial and temporal contrastive patterns.
Furthermore, ST-SSL~\cite{ji2023spatio} utilize self-supervised learning paradigms to capture the spatial and temporal differnece to boost traffic pattern representations.

Besides, the transformer and diffusion architectures are increasingly emerging as powerful methods.
PDFormer~\cite{jiang2023pdformer} employs a self-attention module to capture spatial correlations and graph masking matrices to emphasize spatial dependencies for trajectories prediction.
Efforts to develop domain-agnostic models are exemplified by STEP~\cite{shao2022pre}, which combines spatio-temporal GNNs with a pre-trained transformer. This integration enhances forecasting capabilities by leveraging extensive historical data.
Rather than fitting value functions by averaging all possible returns. TransformerLight~\cite{wu2023transformerlight} utilizes a dynamic system perspective to improve and focuse on producing better actions based on a gated Transformer via substituting residual connections with gated transformer blocks.
Building on the foundation of the transformer, ST-LLM~\cite{liu2024spatial} further harnesses the representation capabilities of large language models. It treats the time steps of a spatial location as tokens and then models the temporal dependencies among these tokens from the global view, specifically to highlight the spatial dimensions, which is tailored to improve prediction accuracy in traffic forecasting.

As for diffusion models, GCRDD~\cite{li2023graph} effectively addresses the challenge of uncertainty quantification in trajectory forecasting by integrating a recurrent structure, while the diffusion process offers a robust and adaptable framework for probabilistic predictions.
Then, KSTDif~\cite{zhou2023towards} employs diffusion models to address the challenge of generating urban flow in data-sparse areas or newly planned regions where historical flow data may be unavailable. This method is particularly useful as it does not rely solely on historical data for predictions.
UniST~\cite{yuan2024unist} is designed to be a universal predictor, addressing the challenge of data scarcity and the need for robust generalization across different urban scenarios.
The use of diffusion within UniST is a strategic choice to enhance the model's generative pre-training capabilities.
Diff-Traj~\cite{zhu2024difftraj} addresses privacy concerns associated with personal location data by incorporating a forward trajectory noising process and a reverse denoising process to effectively simulate the complex, stochastic nature of human movements. 
Furthermore, Diff-RNTraj~\cite{wei2024diff} addresses the limitations of Diff-Traj~\cite{zhu2024difftraj} by ensuring spatial validity and integrating road-specific details without the need for subsequent map-matching, making it more efficient and accurate for practical applications.
TrajGDM~\cite{chu2024simulating} utilizes the diffusion model to capture universal mobility patterns. It represents a significant step forward, using a deep learning network to progressively reduce uncertainty in the trajectory generation process. 
More recently, inspired by the success of large language models, ChatTraffic~\cite{zhang2023chattraffc} generates traffic scenarios from textual descriptions using a diffusion model. This method uniquely combines GCN with diffusion models to incorporate spatial road network structures into traffic predictions to improve the robustness of model to unusual events and improve the model prediction performance on long-term view.

\subsubsection{Time Series}

\textbf{Non-Generative Methods:}  Road-level traffic flow data can be modeled as a time series. Non-generative based methods are shown in details. GWN~\cite{shleifer2019incrementally} combines 1D dilated convolutions and diffusion graph convolutions for accurate traffic prediction. DSANet~\cite{huang2019dsanet} uses CNNs and self-attention for temporal and spatial information capture. Z-GCNETs~\cite{chen2021z} incorporates time-aware zigzag topological layers for Ethereum blockchain price prediction. TAMP~\cite{chen2021tamp} constructs a supragraph convolution module for Ethereum price prediction based on time-conditioned topological properties.

\noindent\textbf{Generative Methods:} More recently,
GraphERT~\cite{beladev2023graphert} introduces a novel method for temporal graph-level embeddings. This approach is pioneering in its use of transformers to seamlessly integrate graph structure learning with temporal analysis.
Besides, DVGNN's~\cite{liang2023dynamic} advantage lies in its ability to consider the interaction between nodes as a stochastic process, which is more aligned with the FMRI dataset, where data is subject to various forms of noise and uncertainty.  
Furthermore, Yu et al.~\cite{yu2023temporal} illustrates the potential of LLMs to provide a unified solution for the challenges of cross-sequence reasoning and inference, as well as the complexities involved in integrating multi-modal signals from historical news.
PromptCast~\cite{xue2023promptcast} directly transforms numerical forecasting into sentence-to-sentence prompts for direct use with LLM.
Time-LLM~\cite{jin2023time} reprograms the input time series using text prototypes prior to feeding it into the frozen LLM, thereby aligning the two modalities.
GPT4MTS~\cite{jia2024gpt4mts} first gathers corresponding textual information, feeding this into the prompt-based LLM framework to simultaneously utilize both temporal data and textual information.
FPT~\cite{zhou2024one} is also inspired by the great success of LLM to fine-tune on all major types of tasks involving time series. 
TEMPO~\cite{cao2023tempo} effectively learns time series representations by focusing on two key inductive biases: decomposing complex interactions between trend, seasonal, and residual components, and designing prompts to facilitate distribution adaptation across various time series types.
LLMTIME~\cite{gruver2024large} demonstrates that LLMs can zero-shot extrapolate time series data at a level comparable to, or even surpassing, that of specialized time series models, without the need for additional text or prompt engineering.

\subsubsection{Spatial Maps}

\textbf{Non-Generative Methods:}  Spatial maps, represented as matrices, are suitable for predictive learning~\cite{DCRNN,zhu2021ast}. ASTGCN~\cite{zhu2021ast} combines STGCN with attention-based graph convolution, improving performance on road sections and considering POI distribution. These methods are non-generative.

\noindent\textbf{Generative Methods:} Recent studies based on diffusion models improves the framework, for instance, DiffSTG~\cite{wen2023diffstg} shows key advantage of using diffusion models in this context is the ability to capture a range of possible future outcomes, which is also validated on air quality monitoring.
USTD's~\cite{hu2023towards} key innovation includes a shared spatial-temporal encoder and task-specific attention-based denoising networks, which allows for accurate air quality monitoring.

\subsubsection{Spatial-Temporal Rasters}

\textbf{Non-Generative Methods:}  Spatial-Temporal raster data is built upon location and time aspects.
ASTGCN~\cite{zhu2021ast} uses spatial-temporal attention mechanisms based on graph convolutional network, proving its robustness on dynamic weather prediction.
GraphCast~\cite{lam2023learning} is a weather forecasting system utilizing graph neural networks to deliver forecasts with a high resolution of 0.25 degrees longitude/latitude, encompassing over a million grid points across the entire Earth's surface.
PastNet~\cite{wu2023pastnet} incorporates a spectral convolution operator in the Fourier domain, which efficiently introduces inductive biases from the underlying physical laws.

\noindent\textbf{Generative Methods:} There are a growing number of works that utilize self-supervision signals.
SVT~\cite{Ranasinghe_2022_CVPR} generates local and global spatial-temporal views with varying spatial dimensions and video frame rates. Its self-supervised objective aims to align the features across these diverse views.
Besides, W-MAE~\cite{man2023w} undergoes self-supervised pre-training to reconstruct spatial correlations among meteorological variables. Subsequently, the model is fine-tuned on a temporal scale to predict future states of these variables, effectively capturing the temporal dependencies inherent in weather data.
LSS-A~\cite{ranasinghe2023language} adapts a pre-trained model for video by modifying its backbone for temporal modeling and training it under self-supervised settings with specific objectives.

More recently, 
SRNDiff~\cite{ling2024srndiff} employs a conditional diffusion framework that integrates historical radar data for more accurate forecasting. 
Hua wt al.~\cite{hua2024weather} introduce a new method for weather forecasting using conditional diffusion models. It emphasizes the integration of numerical weather prediction outputs with diffusion processes, enhancing flexibility and reducing the need for retraining. 
Chen et al.~\cite{chen2023prompt} have been proposed to address data exposure concerns across regions by collaboratively learning a new spatio-temporal Transformer-based foundation model using heterogeneous meteorological data from multiple participants. Additionally, this approach incorporates a unique prompt learning mechanism to meet the communication and computational constraints of low-resourced sensors.

%% file: future.tex
\section{Future Research Directions}
\label{sec:future}
In this section, our objective is to investigate into the future research avenues of spatial-temporal data mining. We present four potential directions, each accompanied by an elaborate description, namely skewed distribution of benchmark datasets, large-scale foundation models, generalization of spatial-temporal methods, and combination with external knowledge.

\noindent\textbf{Skewed Distribution of Benchmark Datasets}.
A skewed distribution in benchmark spatial-temporal datasets refers to an uneven or imbalanced distribution of data points across the spatial and temporal dimensions. This means that certain regions or time periods or different datasets of different tasks may have a significantly higher or lower number of data points compared to others. As a result, the datasets may exhibit a bias towards specific locations or time periods, potentially impacting the accuracy and reliability of any analysis or predictions made using the dataset. More efforts can be made to solve the issue of distributional biases and ensure fair and accurate analyses and models.

\noindent\textbf{Large-Scale Foundation Models}.
Thus far, the limited availability of extensive, high-quality multi-modal datasets has hindered the exploration and advancement of large-scale foundation models. Consequently, there is a pressing need to delve further into the realm of these models to enhance their performance on downstream tasks, particularly in the domain of spatial-temporal forecasting. By addressing this research gap and delving into the development and utilization of large-scale foundation models, we can unlock their potential to significantly enhance accuracy and effectiveness in a wide range of spatial-temporal forecasting applications. 

\noindent\textbf{Generalization of Spatial-Temporal Methods}.
The existing methods in the field of spatial-temporal analysis face challenges when it comes to adapting them to diverse tasks, primarily due to their limited generalization ability. This limitation poses a significant hurdle in utilizing these methods effectively across various domains and scenarios, as they struggle to capture the inherent complexities and nuances present in different tasks. Consequently, there is a crucial need to explore novel approaches that possess enhanced generalization capabilities, enabling seamless adaptation and improved performance across a wide range of spatial-temporal analysis tasks. By addressing this limitation, we can unlock the full potential of spatial-temporal methods and empower researchers and practitioners to tackle diverse challenges in a more flexible and robust manner.

\noindent\textbf{Combination with External Knowledge}.
In light of the ongoing advancements in knowledge graphs, it has become increasingly crucial to explore the integration of external knowledge derived from these graphs as a means to enhance the performance of spatial-temporal methods. The incorporation of such external knowledge holds significant promise in augmenting the analytical capabilities of spatial-temporal methods, enabling them to leverage a broader range of contextual information and domain expertise. By effectively harnessing the wealth of knowledge present within these graphs, researchers and practitioners can unlock new avenues for improving the accuracy, robustness, and overall performance of spatial-temporal methods across diverse applications and domains. Hence, the exploration of methodologies that facilitate the seamless integration of external knowledge from knowledge graphs is of paramount importance in the pursuit of advancing the field of spatial-temporal analysis.

%% file: conclusion.tex
\section{Conclusion}
\label{sec:conclusoin}
In conclusion, this paper sheds light on the integration of \emph{generative techniques} into spatial-temporal data mining, acknowledging the growth and complexity of this data domain. While non-generative techniques like RNNs and CNNs have made significant progress in capturing temporal and spatial dependencies, the emergence of \emph{generative techniques} such as large language models and diffusion models presents new opportunities for advancing spatial-temporal data mining. This paper provides a thorough analysis of spatial-temporal methods based on \emph{generative techniques} and introduces a standardized framework tailored for the data mining pipeline. By offering a comprehensive review and a novel taxonomy, this paper enhances our understanding of the diverse techniques employed in this field. Moreover, it highlights promising future research directions, urging researchers to explore untapped opportunities and push the boundaries of knowledge in order to unlock fresh insights and enhance the effectiveness of spatial-temporal data mining. By integrating \emph{generative techniques} and providing a standardized framework, this paper contributes to the advancement of the field of \emph{generative techniques}.